\documentclass[10pt,twocolumn,letterpaper]{article}

\usepackage[pagenumbers]{wacv}

\usepackage{multirow}   
\usepackage{array}      
\usepackage{siunitx}
\newcommand{\NumScenes}{666}
\newcommand{\NumFrames}{132{,}996}
\newcommand{\NumSampleData}{4{,}654{,}860}
\newcommand{\NumChannels}{35}
\newcommand{\DatasetSizeGB}{2{,}810}
\newcommand{\DatasetSizeTB}{2.74}

\definecolor{wacvblue}{rgb}{0.21,0.49,0.74}
\usepackage[pagebackref,breaklinks,colorlinks,allcolors=wacvblue]{hyperref}
\hypersetup{
  pdftitle={FogDrive: A Multi-Modal Synthetic Driving Dataset for Perception under Graded Fog},
  pdfauthor={Vansh Panwar, Naveen Varshney, Prashant W. Patil}}

\def\confName{WACV}
\def\confYear{2027}

\title{FogDrive: A Multi-Modal Synthetic Driving Dataset for Perception under Graded Fog}

\author{Vansh Panwar \qquad Naveen Varshney \qquad Prashant W. Patil\\
Indian Institute of Technology Guwahati\\
{\tt\small p.vansh@iitg.ac.in}
}

\begin{document}
\maketitle
\addtocontents{toc}{\protect\setcounter{tocdepth}{-1}}
\begin{abstract}
Fog degrades autonomous-driving perception across modalities: it lowers
image contrast, attenuates and scatters LiDAR returns, and shifts the input
distribution of every learned module downstream. Choosing between restoring
the degraded input and training on it directly requires the same scenes observed with and without fog at known
severities, and enough foggy data to train on. Real-world capture cannot
supply this: fog arrives entangled with lighting and traffic, at severities
recorded rather than chosen; existing synthetic datasets narrow the sensor
suite, the scale, or the severity range. We present FogDrive, a CARLA
dataset of $\sim$660 scenes ($\sim$133k frames), each released as a clear
reference plus three calibrated fog densities, with physics-based fog on
camera and LiDAR and 2D/3D annotations consistent across all variants.
Every frame carries four cameras (RGB, depth, semantic), LiDAR, semantic
LiDAR and radar -- spanning fog's response spectrum from heavily degraded
cameras to largely unaffected radar. Experiments trained on FogDrive show
that mixing fog densities improves 3D detection at no extra data cost, that
PSNR and SSIM poorly predict a dehazer's detection benefit, and that radar
improves camera--LiDAR fusion at every density. The dataset and its
generation framework will be released.
\end{abstract}

\section{Introduction}
\label{sec:intro}

Robust perception under adverse weather is a fundamental requirement for autonomous driving, because failures in environmental perception propagate directly into planning and control. Fog is a particularly challenging case: it lowers image contrast through atmospheric scattering, attenuates active sensors and adds spurious backscatter returns, and shifts the input distribution of every learned module that consumes them~\cite{bijelic_2020_stf,sakaridis_2018_ijcv,hahner_2021_lidar_fog}.

Two widely-studied strategies for fog robustness are \emph{defog-then-detect} -- inserting an image-dehazing module~\cite{he_2009_dcp,song_2023_dehazeformer} before a clean-trained detector~\cite{li_2017_aodnet,li_2019_reside} -- and \emph{train-on-fog} -- training the detector directly on degraded observations~\cite{sakaridis_2018_ijcv,bijelic_2020_stf,hahner_2021_lidar_fog}; domain adaptation and image-adaptive detectors are further active directions. The preferrable method and how the trade-off varies with fog density and sensing modality, is difficult to evaluate rigorously without \emph{controlled} environmental variation: matched clean and foggy observations of the same scene at multiple calibrated severities. Real-world foggy datasets cannot supply this; weather cannot be commanded -- fog arrives entangled with changes in lighting, traffic and scene content, at severity that is recorded rather than chosen, and usually without a matched clear counterpart of the same scene. Synthetic datasets can command it, and do provide calibrated, paired fog, but individually narrow the sensor suite, the scale, or the range of severities offered: none simultaneously provides paired clean and degraded observations at multiple calibrated densities, a sensor suite spanning camera, LiDAR and radar, consistent 2D and 3D annotations, and the scale needed to \emph{train} on foggy data rather than only evaluate on it.

We introduce \emph{FogDrive}, a CARLA-based~\cite{dosovitskiy_2017_carla} synthetic multi-modal dataset built to close exactly that gap. Every scene is released in four perfectly aligned variants -- one clear-weather reference and three calibrated fog densities (visibility $\approx$165, 100, and \SI{50}{m}) -- with identical scene geometry and annotations across all four, so perception algorithms can be compared under progressively increasing degradation. FogDrive comprises $\sim$660 scenes ($\sim$133k synchronised frames) across six CARLA towns with a balanced 50:50 day/night split, and provides multi-view RGB with paired depth and semantic segmentation, LiDAR, semantic LiDAR, and radar. Fog is simulated independently for camera and LiDAR using physics-based models -- Koschmieder's law~\cite{hahner_2019_foggy_synscapes} and Beer--Lambert attenuation~\cite{hahner_2021_lidar_fog}, respectively -- and the 2D and 3D annotations are consistent across all weather conditions.

Our major contributions are:
\begin{itemize}
\item We release FogDrive, a large-scale synthetic multi-modal driving dataset of $\sim$660 scenes ($\sim$133k frames) spanning four camera positions with paired depth and semantic segmentation, LiDAR, semantic LiDAR, and radar -- a sensor suite that spans fog's response spectrum -- captured under identical scene geometry across weather conditions.

\item We pair every clear-weather observation with three calibrated fog densities produced by physics-based camera and LiDAR degradation, at a scale that lets \emph{defog-then-detect} and \emph{train-on-fog} be trained and compared directly on the same scenes under increasing degradation.

\item We report experiments on 3D, 2D and radar-fusion detection under graded fog, which yield three findings: training on multiple fog densities beats single-density training; training directly on degraded observations outperforms defog-then-detect, while PSNR and SSIM fail to predict downstream detection performance; and the fog-insensitive radar improves camera--LiDAR fusion at every density.
\end{itemize}

\section{Related work}
\label{sec:related}

Benchmark datasets have driven autonomous-driving perception by standardising evaluation of detection, sensor fusion, and robustness. Datasets' two families: real-world captures, covering both foundational multi-modal benchmarks and adverse-weather collections; and synthetic ones, spanning fog datasets built for image restoration and simulator-generated driving benchmarks. Table~\ref{tab:datasets} summarises how they compare on the properties that matter here.

KITTI~\cite{geiger_2012_kitti} introduced the first widely adopted benchmark combining stereo cameras, LiDAR, and GPS/IMU for detection and localization. nuScenes~\cite{caesar_2020_nuscenes} added a 360\textdegree{} suite (six cameras, one LiDAR, five radars) with large-scale 3D annotations; Waymo Open~\cite{sun_2020_waymo} scaled geographic and annotation density; and BDD100K~\cite{yu_2020_bdd100k} emphasised visual diversity across weather, scene, and time of day. These established the scale and annotation standards of modern perception research; weather, however, is not a variable they set out to control.
A second line of real-world work targets adverse conditions directly.
Foggy Driving~\cite{sakaridis_2018_ijcv} offers 101 real foggy images with semantic labels, used for evaluation rather than training. Seeing-Through-Fog (DENSE/STF)~\cite{bijelic_2020_stf} records $\sim$10\,000\,km with a broad sensor suite (RGB, gated NIR, LiDAR, radar, FIR) and $\sim$100\,k annotations under naturally occurring fog and snow. CADC~\cite{pitropov_2021_cadc} targets snow (7\,k frames). ACDC~\cite{sakaridis_2021_acdc} spans fog, night, rain, and snow (4{,}006 images) with matched clear references, but is camera-only with only $\sim$1{,}000 uncalibrated fog images. These capture authentic degradation, with the caveat inherent to real capture: the weather cannot be dialled in, so severity-matched comparisons across modalities are hard to reproduce.

Foggy Cityscapes~\cite{sakaridis_2018_ijcv} synthesises fog on Cityscapes via Koschmieder's law at three calibrated levels ($\beta = 0.005/0.01/0.02$\,m$^{-1}$; $\approx$600/300/150\,m visibility), and SynFog~\cite{xie_2024_synfog} improves synthetic-fog realism while preserving paired supervision. Both advanced dehazing research considerably; being camera-only, without synchronised LiDAR/radar or 3D annotations, they leave the downstream multi-modal and 3D-detection question open. Multifog KITTI~\cite{mai_2021_multifogkitti} extends the idea to both modalities, fogging KITTI's images and point clouds together over a \SIrange{20}{80}{m} visibility range, and so pairs degraded multi-modal input with the original clear capture; it inherits KITTI's front-facing rig and its scale.

Driving simulators widen the sensor suite.
Early efforts re-created KITTI in simulation -- Virtual KITTI\,2~\cite{cabon_2020_vkitti2} under multiple weathers, KITTI-CARLA~\cite{deschaud_2021_kitticarla} in CARLA (35\,k frames) -- at small scale and aimed at odometry and segmentation. SHIFT~\cite{sun_2022_shift}, the largest CARLA benchmark (2.5\,M+ frames), models weather as a continuous parameter but provides no paired observations across calibrated fog levels. More recently, collaborative-perception datasets have brought controlled fog to several modalities at once: Adver-City~\cite{karvat_2024_advercity} ($\sim$24\,k frames, single fog level) and, most completely, SCOPE~\cite{gamerdinger_2024_scope}, which pairs physically grounded camera and LiDAR fog at three densities and thereby validates the dual-modality fog simulation this paper builds on. Both serve collective perception; at tens of scenarios and tens of thousands of frames, they hold an order of magnitude less data and scenario diversity than the training experiments of \S\ref{sec:experiments} consume.

Arbitrating between defog-then-detect and train-on-fog places three demands
on a dataset that fog realism alone does not meet. Train-on-fog is a
\emph{training} strategy, so the dataset must be large and varied enough to
train detectors and dehazers on foggy data, not only to evaluate them.
The fog must be graded over calibrated severities, so that the comparison
can be resolved by density rather than averaged over it. And because fog
acts along a spectrum of modalities -- degrading the camera most,
attenuating LiDAR, and leaving radar largely untouched
(\S\ref{sec:degradation}) -- the sensor suite should span that spectrum,
radar included, as the natural control arm for modality-robustness
questions. Each line of work above meets some of these demands; none was
built to meet all three. FogDrive is: \NumScenes{} scenarios and
\NumFrames{} frames across six towns, every frame carrying four
RGB\,+\,depth\,+\,semantic camera views, LiDAR, semantic LiDAR and radar,
with 2D/3D annotations held consistent across a clear reference and three
visibility-calibrated fog densities. The experiments of
\S\ref{sec:experiments} exercise all three properties: its scale trains
every detector and dehazer we evaluate, its gradation resolves their
comparison by density, and its radar carries the modality-robustness
question directly.
\begin{table*}[t]
\centering
\footnotesize
\setlength{\tabcolsep}{3pt}
\begin{tabular}{@{}lcllrcccccccc@{}}
\toprule
Dataset & Year & Source & Scenes/frames & Cam & LiD & Rad & SS & 3D & D/N & Fog levels & Ego \\
\midrule
KITTI~\cite{geiger_2012_kitti}              & 2012 & Real      & - / 15\,k         & 4 & \checkmark & - & -        & \checkmark & -        & -                       & \checkmark \\
Foggy Driving~\cite{sakaridis_2018_ijcv}    & 2018 & Real      & - / 101         & 1 & -        & - & \checkmark & -        & -        & real, uncal.~(1)          & \checkmark \\
Foggy Cityscapes~\cite{sakaridis_2018_ijcv} & 2018 & Sim (CS)  & - / 15\,k+      & 1 & -        & - & \checkmark & -        & -        & 3 (600/300/150\,m)        & \checkmark \\
nuScenes~\cite{caesar_2020_nuscenes}        & 2020 & Real      & 1\,k / 1.4\,M     & 6 & \checkmark & \checkmark & -  & \checkmark & \checkmark & -                       & \checkmark \\
Waymo Open~\cite{sun_2020_waymo}            & 2020 & Real      & 1.15\,k / 230\,k  & 5 & \checkmark & - & \checkmark & \checkmark & \checkmark & -                       & \checkmark \\
DENSE/STF~\cite{bijelic_2020_stf}           & 2020 & Real      & - / $\sim$13\,k & 2+g & \checkmark & \checkmark & - & \checkmark & \checkmark & real, uncal.              & \checkmark \\
ACDC~\cite{sakaridis_2021_acdc}             & 2021 & Real      & - / 4\,k$\times$2 & 1 & -        & - & \checkmark & -        & \checkmark & real, uncal.~(1)          & \checkmark \\
SHIFT~\cite{sun_2022_shift}                 & 2022 & Sim (CARLA) & 4.8\,k+ / 2.5\,M+ & 5+ & \checkmark & - & \checkmark & \checkmark & \checkmark & cont.                   & \checkmark \\
SynFog~\cite{xie_2024_synfog}               & 2024 & Sim       & 500 / 5\,k    & 1 & -        & - & \checkmark & -        & \checkmark & 3 (600/300/150\,m)        & \checkmark \\
SCOPE~\cite{gamerdinger_2024_scope}         & 2024 & Sim (CARLA) & 40+ / 17.6\,k   & 1+ & \checkmark & - & \checkmark & \checkmark & \checkmark & 3 (phys.)             & V2X \\
\textbf{FogDrive (ours)}                    & \textbf{2026} & \textbf{Sim (CARLA)} & \textbf{666 / 133\,k} & \textbf{4} & \checkmark+sem. & \checkmark & \checkmark & \checkmark & \checkmark & \textbf{3 ($\approx$165/100/50\,m)} & \checkmark \\
\bottomrule
\end{tabular}
\caption{Representative AD perception datasets. \emph{Cam}: independent RGB
cameras; \emph{LiD/Rad/SS}: LiDAR / radar / pixel-wise semantic segmentation;
\emph{3D}: 3D boxes; \emph{D/N}: day+night; \emph{Fog levels}: distinct
calibrated densities (cont.~= continuous); \emph{Ego}: single-vehicle
(\checkmark) vs.\ collaborative (V2X). ACDC's count is adverse images plus
their clear-weather references; frame counts otherwise follow each dataset's
own convention. Further datasets are discussed in prose but omitted for
length.}
\label{tab:datasets}
\end{table*}

\section{The FogDrive dataset}
\label{sec:dataset}

\subsection{Sensor and vehicle configuration}
\label{sec:sensor_config}

\begin{figure}[t]
  \centering
  \includegraphics[width=0.60\linewidth]{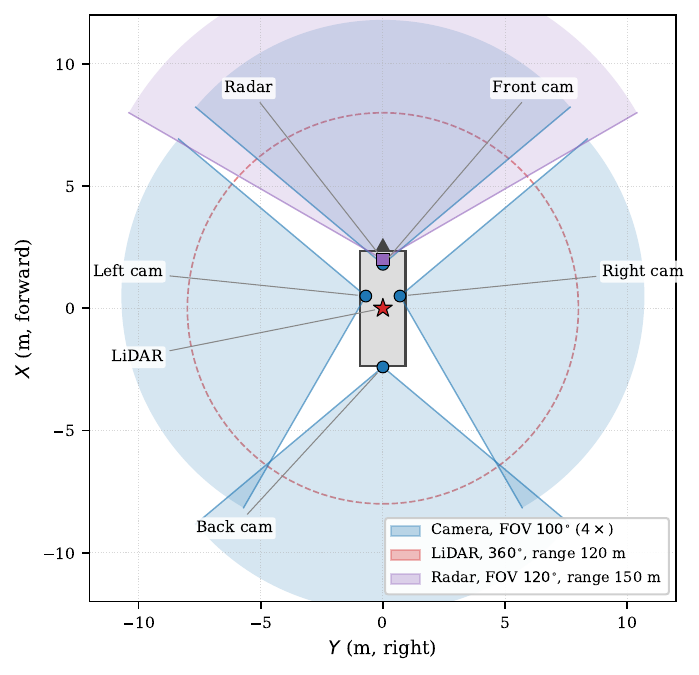}
  \caption{Sensor layout on the ego vehicle (bird's-eye view). FOV cones are
  drawn at a fixed visual length for legibility, not to scale.}
  \label{fig:sensor_layout}
\end{figure}

We use a Tesla Model~3 (blueprint id \texttt{vehicle.tesla.model3}) as the
ego vehicle in every scenario. The rig comprises four RGB cameras, 
each co-located with a depth and
semantic-segmentation camera of identical pose, one roof-mounted
\SI{360}{\degree} LiDAR, one semantic LiDAR co-located with the LiDAR, and
one forward radar. All cameras render at $1280\times960$\,px with a
$100^{\circ}$ horizontal FOV; both LiDARs are 64-channel with a
\SI{120}{m} range, \SI{20}{Hz} rotation and a $[-20,+5]^{\circ}$ vertical
FOV; the radar spans $120^{\circ}\times30^{\circ}$ to \SI{150}{m}. The
layout is shown in Figure~\ref{fig:sensor_layout}; numerical poses
(relative to the ego origin, in CARLA's left-handed $+X$-forward,
$+Y$-right, $+Z$-up frame) and the full per-sensor attribute tables are
given in supplementary \S\,D5. The rig is identical across every
scenario in the dataset.

\subsection{Degradation}
\label{sec:degradation}

\begin{figure}[t]
  \centering
  \includegraphics[width=0.92\linewidth]{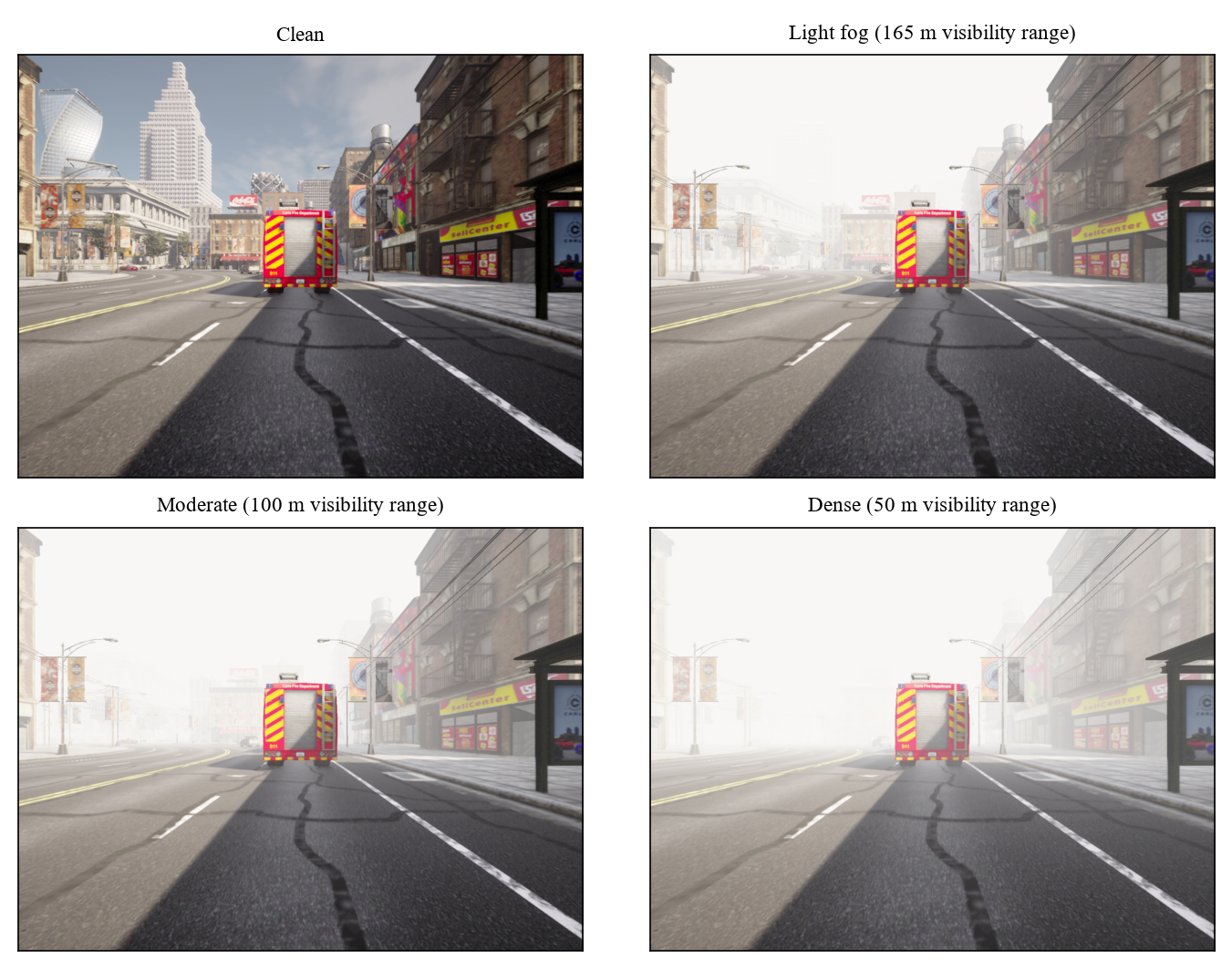}
  \caption{Camera fog at the three calibrated densities (front camera): a
  clean reference and light / moderate / dense fog for one frame.}
  \label{fig:fog_camera}
\end{figure}

\begin{figure}[t]
  \centering
  \includegraphics[width=\linewidth]{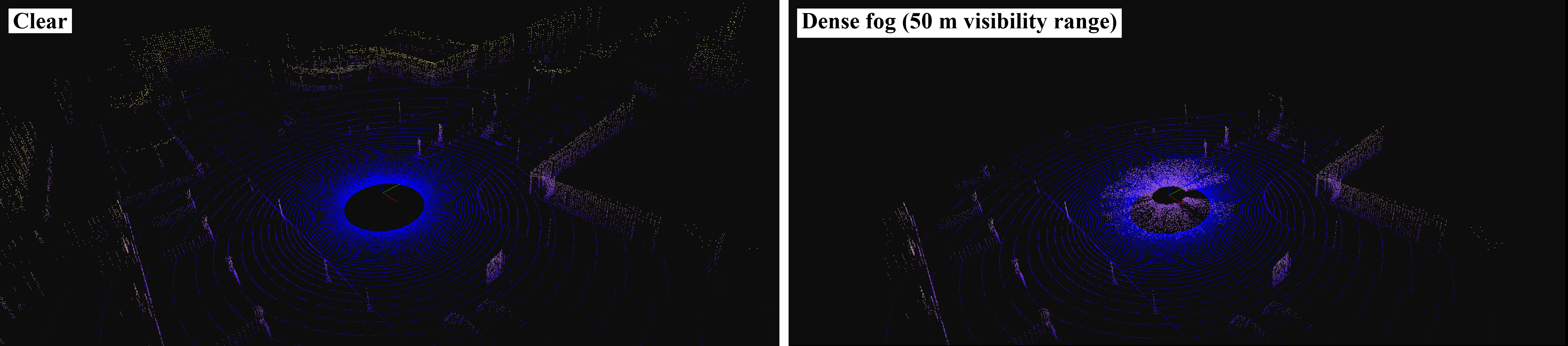}
  \caption{LiDAR fog degradation (Open3D, coloured by height). Dense fog adds
  near-sensor backscatter and collapses the usable range; all four densities
  are shown in supplementary Figure~S3.}
  \label{fig:fog_lidar}
\end{figure}

We release each frame's camera and LiDAR data both clean and at three
calibrated fog densities - \emph{light}, \emph{moderate}, \emph{dense}
- corresponding to meteorological visibility distances of approximately
\SI{165}{m}, \SI{100}{m} and \SI{50}{m} respectively, where visibility
distance is defined as the range at which an object's contrast drops to
\SI{5}{\percent} of its clear-air value (the standard
\SI{5}{\percent}-contrast threshold used in the cited works). Camera fog
is produced with the FoggySynscapes
pipeline~\cite{hahner_2019_foggy_synscapes} (Figure~\ref{fig:fog_camera})
and LiDAR fog with the ETH Zurich LiDAR fog
simulator~\cite{hahner_2021_lidar_fog}. The two tools
share a common physical basis: camera fog applies Koschmieder's atmospheric
optics model per pixel, $I(x) = J(x)\,t(x) + A\,(1 - t(x))$, where $J(x)$
is the clean radiance, $A$ the atmospheric light, and the transmission
$t(x) = \exp(-\beta\,d(x))$ decays exponentially with scene depth $d(x)$
and an extinction coefficient $\beta$ monotone in fog density (depth comes
from CARLA's depth camera). LiDAR fog attenuates each return over its
round trip by the Beer--Lambert law, $P_{\mathrm{rec}} \propto
P_{\mathrm{emit}}\,\exp(-2\alpha R)$ at range $R$; where the fog's
backscattered response exceeds the attenuated hard-target response the
return is replaced by a nearer scattering point, and otherwise only its
intensity falls (Figure~\ref{fig:fog_lidar}). The attenuation coefficient
$\alpha$ of Table~\ref{tab:fog_config} plays the role of $\beta$ above;
$\alpha = 0.06, 0.03, 0.018$ give $V \approx \ln(20)/\alpha = 50, 100,
165$\,m at the \SI{5}{\percent}-contrast threshold.

\begin{table}[t]
\centering
\scriptsize
\setlength{\tabcolsep}{4pt}
\begin{tabular}{@{}lcccc@{}}
\toprule
Fog level & $\alpha$ (extinction) & $\gamma$ (backscatter) & Noise scale & Noise variant \\
\midrule
Dense    & 0.060 & $10^{-6}$ & 10 & v2 \\
Moderate & 0.030 & $10^{-6}$ & 10 & v2 \\
Light    & 0.018 & $10^{-6}$ & 10 & v2 \\
\bottomrule
\end{tabular}
\caption{Fog-simulator configuration. $\alpha$: extinction (larger
$\Rightarrow$ shorter visibility); $\gamma$: residual backscatter; noise:
additive per-point droplet scatter. LiDAR intensities are stored on CARLA's
native \mbox{0--1} scale.}
\label{tab:fog_config}
\end{table}

\paragraph{Radar is released without fog degradation.}
We deliberately leave the radar channel undegraded: a \SI{77}{GHz} radar's
${\sim}\SI{4}{mm}$ wavelength is orders of magnitude larger than fog
droplets (${\sim}$1--\SI{40}{\micro\metre}), placing it deep in the
Rayleigh regime where extinction is negligible, whereas LiDAR
(905/\SI{1550}{nm}) sits in the Mie regime and is strongly
attenuated~\cite{skolnik_2001_radar,itu_2019_p840,srivastav_2023_radar,bijelic_2020_stf,elhaddioui_2025_montecarlo}.
A synthetic radar-fog model would therefore inject an artefact rather than
reproduce a real failure mode; the supporting measurements are tabulated in
supplementary Table~S6.

\subsection{Data format and annotation}
\label{sec:format}

Scenes are 20-second segments of synchronised multi-sensor data captured
at \SI{20}{Hz} simulation rate, with sensor outputs and annotations saved
every \SI{0.1}{s} (\SI{10}{Hz}; $\sim$200 keyframes per scene). The
annotations were generated by modifying the open-source CARLA-2DBBox community
module by Adib~\cite{adib_2020_carla_2dbbox}; the released images have
been stripped of CARLA's alpha channel.

Vehicles are annotated from the semantic LiDAR: an actor returning at least
\texttt{min\_detect}~$= 5$ semantic-LiDAR points within
\texttt{max\_dist}~$= \SI{100}{m}$ receives a 3D box (all eight corners, in
the LiDAR frame), which is then projected into each camera with the
per-scene extrinsic and the shared intrinsic to give its axis-aligned 2D
box; the same visibility test drives both. Our multi-camera extension of the
open-source annotator -- sharing one semantic LiDAR across all four cameras
in a single scene sweep -- and its geometric limitation (cameras must stay
near the LiDAR) are described in supplementary \S\,D3;
\S\ref{sec:validation} quantifies the effect on the released annotations.

FogDrive annotates three vehicle classes,
hand-curated per CARLA blueprint:
$\{0 = \text{passenger car}, 1 = \text{truck (incl.\ buses and fire
  trucks)}, 4 = \text{bigger car (vans, large SUVs)}\}$. The
ids are inherited from the broader source taxonomy -- hence their
non-contiguity. The
full 37-entry blueprint-to-class mapping is in supplementary \S\,D1.

\begin{table}[t]
\centering
\footnotesize
\setlength{\tabcolsep}{6pt}
\begin{tabular}{@{}lc@{}}
\toprule
Channel family & Variants per camera \\
\midrule
RGB camera (clean)     & 4 (front/back/left/right) \\
RGB camera (foggy)     & $4 \times 3$ levels \\
Depth camera           & 4 \\
Semantic camera        & 4 \\
2D bbox file           & 4 \\
LiDAR (clean)          & 1 \\
LiDAR (foggy)          & 3 levels \\
Semantic LiDAR         & 1 \\
Radar                  & 1 \\
3D bbox file           & 1 \\
\bottomrule
\end{tabular}
\caption{Sensor channels released per sample (35 total).}
\label{tab:channels}
\end{table}

\subsection{Accessing the dataset}
\label{sec:access}

We release four nuScenes-style metadata tables
(\texttt{scene}, \texttt{sample}, \texttt{sample\_data}, \texttt{splits})
and a \texttt{FogDrive} Python access class that loads them into RAM and
answers the common ``give me channel $c$ for sample $s$'' queries in a single
lookup, together with a reference PyTorch dataset wrapper. The full JSON
schema (supplementary \S\,D6), access-class methods (\S\,D4), and on-disk
directory tree (\S\,D2) are given in the supplementary material.

\subsection{Scenario diversity and generation strategy}
\label{sec:diversity}

FogDrive is collected in CARLA~\cite{dosovitskiy_2017_carla} using a single
ego rig and CARLA's built-in Traffic Manager to drive both the ego vehicle
and the background traffic. The collection script is configurable in
town, time-of-day, scenario length, sensor configuration and NPC density,
and was re-run with different settings to populate the dataset; this
subsection documents the configuration that produced the released data.

A scenario is a single 20-second segment captured at \SI{20}{Hz}, with
keyframes annotated and saved every \SI{0.1}{s} (\SI{10}{Hz}; $\sim$200
samples per scenario). Each scenario is independent: at its start the world
is reseeded with a fresh ego, background NPCs, and sensor rig, all destroyed
at its end. Scenarios span six CARLA towns (Town01--05, Town10HD) in two
time-of-day states -- ``day'' (CARLA's default preset) and ``night''
(sun altitude $-90^{\circ}$). Collection-time details (map
``optional-layers'' handling, traffic-light phasing) are given in
supplementary \S\,D9.

The ego is the Tesla Model~3, spawned at a random valid spawn-point and
driven by CARLA's Traffic Manager; trajectory diversity arises from the
random spawn, the random per-scenario background traffic, and the Traffic
Manager's stochastic intersection and lane-change behaviour. Background
traffic is drawn from the full CARLA \texttt{vehicle.*} pool minus the
seven two-wheeler blueprints, with a town-dependent NPC cap
(supplementary \S\,D9). FogDrive therefore annotates four-wheeled vehicles
only; pedestrians are out of scope, in line with the tasks in
\S\ref{sec:experiments}.

All scenes are collected in CARLA's synchronous mode, so every released
sample is a coherent snapshot of all 15 channels at one simulator
timestamp, with no inter-sensor interpolation required downstream. All
sensor frames follow CARLA's left-handed convention ($+X$ forward,
$+Y$ right, $+Z$ up); users composing extrinsics under right-handed
conventions should apply the corresponding handedness flip. The radar's
spherical-coordinate storage and Cartesian recovery are given in
supplementary \S\,D8. All modalities are released in lossless on-disk
formats; the per-sensor specification -- array shapes, channel
conventions, the depth encoding, and the bounding-box JSON schemas --
is given in supplementary \S\,D8 and \S\,D6.

\subsection{Dataset statistics}
\label{sec:statistics}

FogDrive contains \NumScenes{} scenes comprising \NumFrames{} synchronised
multi-modal frames captured at \SI{10}{Hz} across six CARLA towns. Each
frame bundles the \NumChannels{} sensor channels of
Table~\ref{tab:channels}, bringing the total sample-data record count to
\NumSampleData{}. The dataset occupies $\DatasetSizeGB$\,GiB
($\approx \DatasetSizeTB$\,TiB) on disk, with camera channels accounting
for $\approx\SI{86}{\percent}$ of the total size (clean RGB alone is
\SI{611}{GB}; the three foggy RGB variants together add a further
\SI{1410}{GB}). The full per-modality breakdown is reproduced in
supplementary \S\,D7. Each scene is a contiguous
\SI{10}{Hz}, \SI{20}{s} capture of $\sim$200 samples and can be treated as
fixed-length.

Scenes are distributed across Town01--05 and Town10HD with budgets
scaled to map size (for example, Town03 carrying 140 scenes, while Town01 accounting for 94 scenes), following the heuristic that larger maps provide more varied geometries to explore. Each town contributes day and night runs in exactly
equal proportion, giving an overall day/night split of \emph{333/333
scenes (50.0\%/50.0\%)}. Scenes are partitioned 70/15/15 into
train/val/test with the 50/50 day/night ratio preserved exactly and
per-town shares matched within $\pm 2$ percentage points in every split
(full split table in supplementary \S\,D6).

Across all \NumFrames{} front-camera frames we observe 300\,604 annotated
vehicle instances, distributed as
\emph{passenger car \SI{64.4}{\percent} (193\,683)},
\emph{truck \SI{14.6}{\percent} (43\,754)} and
\emph{bigger car \SI{21.0}{\percent} (63\,167)}.
The mean number of annotated vehicles per front-camera frame is
2.26 (median~2, p25/p75~$=1/3$, max~15);
\emph{\SI{21.0}{\percent} of frames contain zero annotated vehicles},
a direct consequence of including night-time scenes on lightly-trafficked
roads and outer-town segments. Detectors trained on FogDrive must handle
empty frames gracefully - a point we return to in
\S\ref{sec:experiments}.

\subsection{Annotation-quality validation}
\label{sec:validation}

The 3D box extents come from CARLA and are correct by construction, so
validation targets the one component we author: the semantic-LiDAR
visibility filter (\S\ref{sec:format}). We audit it against a reference
that shares none of its machinery -- CARLA's semantic-segmentation
camera, which labels each pixel directly from scene metadata rather than
from LiDAR returns, and is therefore independent of the filter under
test. Over a stratified sample of 8\,096 images (506 scenes, 12
town\,$\times$\,time-of-day strata, all four cameras) we build a binary
vehicle-pixel mask and score the released annotations against it at
pre-registered coverage thresholds, counting a vehicle as recalled when
an annotated box covers most of its pixels, and a box as correct when
enough of the pixels it encloses are vehicle-class.

The filter is effectively exact within the LiDAR's return range and
degrades gracefully beyond it. Recall is $\geq\SI{99}{\percent}$ within
${\approx}\SI{40}{m}$ and \SI{87.8}{\percent} overall, the shortfall
falling almost entirely on distant vehicles that return too few LiDAR
points to trigger the filter rather than on annotation error; precision
is \SI{95.1}{\percent} (\SI{94.3}{\percent} under the strictest
accounting). A sensitivity sweep confirms neither headline number is an
artefact of one particular threshold. The full protocol, the
recall-by-distance and per-camera breakdowns, and the failure-mode
taxonomy with examples are given in supplementary \S\,D11; the audit
script and pre-registered protocol are released with the dataset.

\section{Experiments}
\label{sec:experiments}

We use FogDrive to investigate three questions that no prior benchmark
supports together at autonomous-driving scale. \emph{Experiment 1} (\S\ref{sec:exp1}) asks whether training
a 3D LiDAR / camera-LiDAR detector on a single fog density generalises
across fog densities, and whether mixing densities at training time helps.
\emph{Experiment 2} (\S\ref{sec:exp2}) examines the 2D image-detection
counterpart: does dehazing the camera input restore detection accuracy,
and does standard reference-based image-quality (PSNR, SSIM) predict
which dehazer will help downstream detection? \emph{Experiment 3}
(\S\ref{sec:exp3}) turns to the modality spectrum: does the fog-insensitive
radar add value to camera--LiDAR fusion under fog? All experiments use the
70/15/15 scene-level splits of \S\ref{sec:statistics}; all test sets are
disjoint at the scene level so clean and foggy variants of every test
scene appear in the same split.

\subsection{Experiment 1: single-density vs.\ mixed-density 3D detection}
\label{sec:exp1}

\paragraph{Setup.}
We compare four training subsets, all drawn from the FogDrive training
split: \textbf{Light} (\SI{20}{\percent} of light-fog scenes),
\textbf{Dense} (\SI{20}{\percent} of dense-fog scenes), \textbf{Mixed-S}
(\SI{8}{\percent} each of light/moderate/dense, $\approx \SI{24}{\percent}$
total), and \textbf{Mixed-L} (\SI{20}{\percent} each of
light/moderate/dense, \SI{60}{\percent} total). The dual design separates
two questions: Mixed-S vs.\ \{Light, Dense\} asks whether diversifying fog
density helps \emph{at a fixed total budget}, and Mixed-L vs.\ Mixed-S
whether \emph{more data per density} helps at fixed diversity. Each model
is evaluated on three held-out test subsets - \SI{20}{\percent} of each
fog density's test scenes - reported independently per test density.

We use TransFusion-L~\cite{bai_2022_transfusion}, the LiDAR-only variant of
TransFusion, and
BEVFusion~\cite{liu_2023_bevfusion} (camera + LiDAR fusion) as
representative 3D detectors. The point-cloud range is $\pm\SI{40}{m}$:
dense-fog LiDAR returns essentially nothing beyond ${\sim}\SI{36}{m}$, so a
wider grid would be mostly empty at that density. For each
(training~$\times$~test) pair we report the checkpoint that maximises the
composite FogDrive Detection Score
($\mathrm{FDS} = 0.5\,\mathrm{mAP} + 0.5\,\mathrm{Mean\,TP}$) on that test
subset; training schedules and optimiser settings are in supplementary
\S\,D10.

We report \textbf{mAP@0.5} (3D IoU=0.5) for detection accuracy,
\textbf{Mean~TP} (the nuScenes-style mean of the three true-positive
error scores, $1 - \text{error}/\text{cap}$, on a 0--1 scale shared with
mAP) for overall localization quality, and \textbf{ATE} (Average
Translation Error, $1 - \text{error}/\SI{2}{m}$) for box position. Its
companions ASE (scale) and AOE (orientation) follow the same direction and
are reported in supplementary \S\,D10. \textbf{Results.}
Table~\ref{tab:exp1-main} reports mAP@0.5 and Mean TP as 4$\times$3
(training~$\times$~test density) matrices per detector, each cell at the
best-FDS checkpoint (ATE and the scale/orientation components are in
supplementary \S\,D10).

\begin{table*}[t]
\centering
\footnotesize
\setlength{\tabcolsep}{4pt}
\begin{tabular}{@{}lcccccc|cccccc@{}}
\toprule
 & \multicolumn{6}{c|}{TransFusion-L} & \multicolumn{6}{c}{BEVFusion} \\
 & \multicolumn{3}{c}{mAP@0.5} & \multicolumn{3}{c|}{Mean TP} & \multicolumn{3}{c}{mAP@0.5} & \multicolumn{3}{c}{Mean TP} \\
Train$\downarrow$ / Test$\rightarrow$ & L & M & D & L & M & D & L & M & D & L & M & D \\
\midrule
Mixed-L  & 0.9158 & \textbf{0.8917} & 0.7582 & \textbf{0.9470} & \textbf{0.9427} & \textbf{0.9420} & \textbf{0.9077} & 0.8565 & 0.7242 & \textbf{0.9560} & \textbf{0.9507} & \textbf{0.9562} \\
Light    & 0.9129 & 0.8881 & 0.7550 & 0.9393 & 0.9280 & 0.9362 & 0.8794 & 0.8433 & 0.7608 & 0.9421 & 0.9310 & 0.9420 \\
Mixed-S  & \textbf{0.9280} & 0.8742 & \textbf{0.7833} & 0.9383 & 0.9348 & 0.9417 & 0.8931 & \textbf{0.8646} & 0.7501 & 0.9410 & 0.9340 & 0.9414 \\
Dense    & 0.8964 & 0.8454 & 0.7640 & 0.9146 & 0.9140 & 0.9384 & 0.8539 & 0.8323 & \textbf{0.7761} & 0.8839 & 0.8872 & 0.9257 \\
\bottomrule
\end{tabular}
\caption{Experiment~1: \textbf{mAP@0.5} and \textbf{Mean TP} as 4$\times$3
(training~$\times$~test) matrices per detector. \emph{L/M/D}: light/moderate/dense
test fog; higher is better, column winner \textbf{bold}, at the best-FDS
checkpoint. ATE and scale/orientation components in supplementary \S\,D10.}
\label{tab:exp1-main}
\end{table*}

\emph{Detection rate scales with diversity, not volume.}
A mixed-density configuration wins five of the six mAP columns across the two
detectors, and single-density Light -- the intuitive choice if one expects
mild fog at test time -- wins none of them. The exception is instructive:
under dense fog on BEVFusion, matched single-density Dense training is best,
which is where the domain gap is widest and the LiDAR sparsest, so training
exactly on the target condition pays. Mixing is otherwise the stronger
default, and it is \emph{diversity} rather than volume that delivers it:
Mixed-S is matched to the single-density subsets on total scene budget, yet
wins the light and dense columns on TransFusion-L and the moderate column on
BEVFusion. Tripling the per-density data in Mixed-L buys a win only where
returns are still plentiful enough to exploit -- it leads at moderate fog on
TransFusion-L and at light fog on BEVFusion, but falls to last at dense fog on
BEVFusion. This is consistent with detection rate being volume-insensitive at
the extremes: light fog is near-clean, so detection already sits near its
ceiling, while dense-fog detection is capped by LiDAR return sparsity (few
points survive beyond ${\sim}\SI{36}{m}$), a physical limit extra training
data cannot lift.

\emph{Scaling the mix tightens localization.}
Once a box is detected, its geometric accuracy is best under the full Mixed-L
mixture: Mixed-L wins/ties Mean TP across all three test densities on
\textbf{both} detectors, with the effect more pronounced on BEVFusion --
plausibly because its camera branch sees different contrast statistics across
densities. The effect is strongest in translation: on TransFusion-L, Mixed-L improves ATE over
Light-trained by up to $+0.019$ (moderate test), $\sim\SI{4}{cm}$ at the
standard \SI{2}{m} cap, and stays essentially flat across test densities
where Light-trained models degrade -- a domain-shift signature in the
translation channel. The ATE matrices and the scale/orientation components,
which show the same direction, are in supplementary \S\,D10.
Mix fog densities at training time whenever possible -- on both detectors it
improves detection rate at no cost in training data, the one exception being
dense fog on BEVFusion, where matched single-density training wins -- and
scale the mix when downstream consumers (trajectory prediction, motion
planning) need accurate box geometry.

\subsection{Experiment 2: defog-then-detect vs.\ train-on-fog (2D)}
\label{sec:exp2}

\begin{figure}[t]
  \centering
  \includegraphics[width=\linewidth]{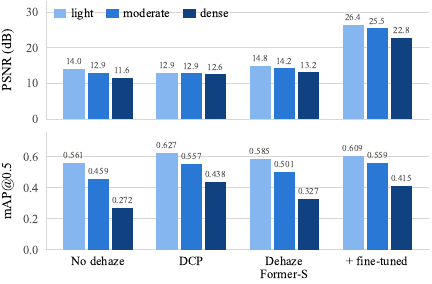}
  \caption{Reconstruction quality and detection utility order the dehazers
  differently. Finetuned DehazeFormer-S by far the most faithful
  reconstructor, but not the best detector input.}
  \label{fig:exp2-pipeline}
\end{figure}

\paragraph{Setup.}
We compare four end-to-end 2D-detection pipelines on the front and left
RGB streams, holding the detector architecture and hyperparameters
constant so that any difference is attributable to the input
transformation rather than to the detector. Pipeline \textbf{A}
(clean$\to$clean) is the clean-trained reference; \textbf{B} (clean$\to$fog) is the
degradation baseline; \textbf{C} (fog$\to$dehazer$\to$detector) is
defog-then-detect with three dehazer configurations; \textbf{D}
(fog$\to$fog) trains a separate detector on each fog level. The
detector is YOLOv8-m~\cite{jocher_2023_yolov8}, initialised from
COCO-pretrained weights and fine-tuned end-to-end (training and validation
settings in supplementary \S\,D12). To keep the experiment tractable we
subsample scenes at $15/20/\SI{20}{\percent}$ (train/val/test); the test set
is 7\,188 frames per fog level across the two cameras. Image quality
(PSNR, SSIM) is computed against the paired clean frame. Pipeline C's three
dehazer configurations span the methodological range and probe the role of
in-domain adaptation: \textbf{DCP}~\cite{he_2009_dcp} (2009, parameter-free
Dark Channel Prior, with guided-filter~\cite{he_2010_guidedfilter}
refinement),
\textbf{DehazeFormer-S}~\cite{song_2023_dehazeformer} (2023,
$\sim$1.3M-parameter Swin-style transformer, authors' RESIDE-OUT weights),
and \textbf{DehazeFormer-S fine-tuned}, the same transformer fine-tuned on
FogDrive paired (clean, foggy) training data from that initialisation. The
fine-tuned variant is the domain-adaptation control: if adapting a modern
transformer to the target domain does not lift its detection above a
parameter-free prior, the image-quality/detection decoupling we observe is
not a domain-transfer artefact. \textbf{Results.}
Table~\ref{tab:exp2-detection} reports detection metrics for every
pipeline / fog level combination on the 7\,188-frame test split, and
Figure~\ref{fig:exp2-pipeline} visualises the headline mAP@0.5 column.
Two findings dominate the table.

\emph{(i) Fog-trained detection matches the clean-trained reference.}
Evaluated on foggy images the clean-trained detector (B) loses
\SIrange{19}{61}{\percent} of its clean mAP@0.5. Training on the degraded
data instead (D) recovers essentially all of it: D matches A at light fog,
exceeds it at moderate, and falls modestly short only at dense. That a
fog-trained detector can equal or beat the clean-trained one follows from
the annotation geometry. FogDrive annotates vehicles to \SI{100}{m}, so
while visibility stays at or beyond that range (light
${\approx}\SI{165}{m}$, moderate ${\approx}\SI{100}{m}$) no annotated
vehicle is lost, yet fog compresses background texture, colour and
illumination toward the atmospheric light, leaving a matched-domain task
statistically \emph{cleaner} than the clean-weather one. This margin
accordingly grows from light to moderate, then reverses at dense, where
visibility (${\approx}\SI{50}{m}$) cuts inside the annotation range and the
\SIrange{50}{100}{m} band becomes unresolvable. The gain is recall:
at dense fog D more than doubles B's recall while also improving precision.

\emph{(ii) Defogging helps, but hits a ceiling well below training on fog.}
Every dehazer improves on the foggy baseline, and each improves more as fog
thickens. The ceiling, however, is low: across the three densities the best
dehazer recovers only a third to a half of the detection lost to fog,
whereas Pipeline~D recovers essentially all of it. No Pipeline~C variant
matches D at any density, and the shortfall widens as fog thickens.
In-domain fine-tuning is the most effective intervention available to a
dehazer -- it lifts DehazeFormer-S at every density, and by progressively
more as fog thickens -- but it is not enough to change that verdict: the
fine-tuned transformer only draws level with DCP, a parameter-free method
from 2009 that uses no training data at all.
\begin{table}[t]
\centering
\scriptsize
\setlength{\tabcolsep}{2.4pt}
\begin{tabular}{@{}llrrrrr@{}}
\toprule
Pipeline & Fog & PSNR & SSIM & mAP@0.5 & P & R \\
\midrule
A (clean$\to$clean) & clean & -- & -- & 0.692 & 0.808 & 0.623 \\
\midrule
B (clean$\to$fog) & light & 14.00 & 0.746 & 0.561 & 0.818 & 0.558 \\
B & moderate & 12.92 & 0.712 & 0.459 & 0.775 & 0.440 \\
B & dense & 11.58 & 0.662 & 0.272 & 0.689 & 0.253 \\
\midrule
C -- DCP & light & 12.95 & 0.648 & 0.627 & 0.794 & 0.606 \\
C -- DCP & moderate & 12.87 & 0.639 & 0.557 & \textbf{0.812} & 0.555 \\
C -- DCP & dense & 12.56 & 0.623 & 0.438 & 0.772 & 0.435 \\
C -- DehazeFormer-S & light & 14.84 & 0.761 & 0.585 & \textbf{0.831} & 0.571 \\
C -- DehazeFormer-S & moderate & 14.25 & 0.735 & 0.501 & 0.796 & 0.490 \\
C -- DehazeFormer-S & dense & 13.15 & 0.692 & 0.327 & 0.779 & 0.312 \\
C -- DehazeFormer-S (ft.) & light & \textbf{26.43} & \textbf{0.890} & 0.609 & 0.800 & 0.595 \\
C -- DehazeFormer-S (ft.) & moderate & \textbf{25.46} & \textbf{0.860} & 0.559 & 0.793 & 0.532 \\
C -- DehazeFormer-S (ft.) & dense & \textbf{22.78} & \textbf{0.805} & 0.415 & 0.771 & 0.408 \\
\midrule
\textbf{D (fog$\to$fog)} & light & -- & -- & \textbf{0.693} & 0.826 & \textbf{0.638} \\
\textbf{D} & moderate & -- & -- & \textbf{0.716} & 0.773 & \textbf{0.673} \\
\textbf{D} & dense & -- & -- & \textbf{0.647} & \textbf{0.851} & \textbf{0.591} \\
\bottomrule
\end{tabular}
\caption{Experiment~2 per fog level: input image quality (PSNR dB, SSIM,
against the paired clean frame) alongside detection (mAP@0.5, precision P,
recall R). A consumes the clean image and D the unprocessed foggy image
(B's input), so IQ is shown for the B and C rows. \textbf{Bold}: best per
fog level among comparable rows. Full per-class metrics in
supplementary \S\,D12.}
\label{tab:exp2-detection}
\end{table}
Reading Table~\ref{tab:exp2-detection}'s image-quality and detection
columns against each other:
\textbf{the two rankings disagree, and the disagreement survives domain
adaptation.} Fine-tuning makes DehazeFormer-S by far the most faithful
reconstructor we test -- it leads DCP by \SIrange{10}{13}{dB} PSNR and
\SIrange{0.18}{0.24}{} SSIM at every density -- and buys it no detection
advantage: it trails DCP at light and dense fog and merely matches it at
moderate. DCP is the \emph{worst} method tested by SSIM, scoring below the
foggy input it is given, and the best or joint-best by detection at every
density. Ordering the three dehazers by reconstruction quality gives a
strictly different order than ordering them by detection.

PSNR and SSIM reward pixel-level similarity to a clean reference --
faithful texture, colour, and fine detail -- which is what a network
minimising L1/L2 reconstruction optimises. But the detector consumes a
feature map, and the cues that matter for vehicle detection (coarse
silhouettes, contrast, edge sharpness) are not what PSNR / SSIM are most
sensitive to. DCP does the opposite of perceptual reconstruction: it
aggressively boosts contrast and removes atmospheric light, producing
outputs far from the clean reference but rich in the high-contrast structure
a convolutional detector relies on. The features the detector relies
on are not the ones PSNR and SSIM measure. This decoupling is consistent
with earlier task-driven evidence - dehazing does not reliably improve
CNN-based classification~\cite{pei_2018_dehaze_classification}, and
RESIDE's detection-based evaluation already ranked dehazers differently
from PSNR and SSIM~\cite{li_2019_reside}; our contribution is to show
that it persists under calibrated, paired multi-density fog and survives
in-domain fine-tuning.

The same reasoning explains the ceiling in finding~(ii). Fog does not merely
distort the image, it destroys information: returns from distant,
low-contrast vehicles are attenuated below what the sensor resolves. A
dehazer can restore the appearance of whatever survived attenuation, but it
cannot recover what was erased -- so even a near-faithful reconstruction
leaves most of the detection gap open. Training on fog wins because it teaches 
the detector to exploit the faint evidence that remains, rather than attempting recovery.
Additionally the task metrics play a better role on the choice of dehazor, 
than just the image quality metrics.

\subsection{Experiment 3: does radar add value under fog?}
\label{sec:exp3}

Radar is the one FogDrive modality released undegraded
(\S\ref{sec:degradation}), which makes it a natural candidate for recovering
what fog removes. We ablate it as a third input stream on the
camera\,+\,LiDAR detector: \textbf{C+L} and \textbf{C+L+R} differ by the
radar branch alone. Radar points are encoded with a pillar-style
scatter to the shared BEV grid~\cite{lang_2019_pointpillars} and fused by
concatenation with the camera and LiDAR streams in BEVFusion's convolutional
fuser~\cite{liu_2023_bevfusion}. Each point carries $[x, y, z, v_r, r]$;
CARLA simulates no radar cross-section, so no RCS channel is available. We train C+L for 8 epochs, freeze it and
train the radar branch and fuser for 5 epochs from its best checkpoint, then
fine-tune the whole model for 8 epochs. Both arms are trained once on
mixed-fog data and evaluated at each test density.

\emph{Radar helps, but not more as fog thickens.} Adding radar improves
mAP@0.5 at every density (Table~\ref{tab:exp3-radar}). The gain does not, however, grow with degradation: it is largest
at moderate fog and smallest at dense, so on this evidence radar is
supplying generic supervision rather than fog-invariance specifically. Two
properties of the release plausibly bound the effect: the radar is
front-facing, covering a $120^{\circ}$ wedge that contains only
\SI{43.4}{\percent} of in-range ground-truth boxes, and it carries no radar
cross-section, a channel radar-fusion methods rely on heavily.

\begin{table}[t]
\centering
\footnotesize
\setlength{\tabcolsep}{7pt}
\begin{tabular}{@{}lccc@{}}
\toprule
Model & Light & Mod. & Dense \\
\midrule
C+L (baseline) & 0.7683 & 0.7609 & 0.6547 \\
C+L+R (radar)  & \textbf{0.7779} & \textbf{0.7734} & \textbf{0.6631} \\
\bottomrule
\end{tabular}
\caption{Experiment~3: radar ablation, mAP@0.5 per test fog density. The two
arms differ by the radar branch alone.}
\label{tab:exp3-radar}
\end{table}

\section{Limitations and Future Work}
\label{sec:limitations}

FogDrive is synthetic, so standard sim-to-real caveats apply: any benchmark
result obtained on it should be validated on real measurements before being
transferred to real fog. The release is also deliberately scoped, and the
extended release we are preparing broadens it along most of these axes.
The present version covers three sizes of four-wheeled vehicles, with ego and background traffic driven by
CARLA's Traffic Manager for lane-aware, uniformly-policied behaviour and a
fixed sensor rig; pedestrian \& two-wheeler annotations, a wider sensor and
modality suite, and a controlled array of crash and accident scenarios of
varied types are in preparation. Degradation is modelled for the camera and
LiDAR channels at 3 calibrated densities, with radar left undegraded by
design; rain and snow, at the same calibrated
multi-intensity granularity as the fog here, are the natural next axis and
would move FogDrive from a foggy benchmark toward a general, controlled
adverse-weather one. Roughly one front-camera frame in five contains no annotation (\S\ref{sec:statistics}), a consequence of covering
low-traffic conditions, so detection pipelines should handle negative frames
gracefully. Residual visibility-filter artefacts are quantified in
\S\ref{sec:validation} and supplementary \S\,D11.

\section{Conclusion}
\label{sec:conclusion}

We presented \emph{FogDrive}, a multi-modal CARLA dataset of $\sim$660
scenes ($\sim$133k frames) paired across a clear reference and three
calibrated fog densities, with annotations consistent throughout.
Experiments on that structure show that in 3D
detection, mixing fog densities at training time improves detection rate
at no extra data cost, and \emph{scaling} the mix further tightens box
geometry; that in 2D detection, training directly on foggy data
matches the clean-trained reference and beats defog-then-detect, while
PSNR/SSIM do not predict detection utility; and that adding the
undegraded radar channel to camera--LiDAR fusion improves detection at
every fog density. The dataset will be released.

{
    \small
    \bibliographystyle{ieeenat_fullname}
    \bibliography{main}
}

\clearpage

\setcounter{section}{0}
\setcounter{table}{0}
\setcounter{figure}{0}
\setcounter{equation}{0}
\renewcommand{\thesection}{D\arabic{section}}
\renewcommand{\thesubsection}{\thesection.\arabic{subsection}}
\renewcommand{\thetable}{S\arabic{table}}
\renewcommand{\thefigure}{S\arabic{figure}}
\renewcommand{\theequation}{S\arabic{equation}}
\def\theHsection{supp.\arabic{section}}
\def\theHsubsection{\theHsection.\arabic{subsection}}
\def\theHtable{supp.\arabic{table}}
\def\theHfigure{supp.\arabic{figure}}
\def\theHequation{supp.\arabic{equation}}

\makeatletter
\renewcommand{\numberline}[1]{#1\hspace{1em}}
\renewcommand*\l@section[2]{%
  \ifnum \c@tocdepth >\z@
    \addpenalty\@secpenalty
    \addvspace{1.0em \@plus\p@}%
    \setlength\@tempdima{2.4em}%
    \begingroup
      \parindent \z@ \rightskip \@pnumwidth
      \parfillskip -\@pnumwidth
      \leavevmode \bfseries
      \advance\leftskip\@tempdima
      \hskip -\leftskip
      #1\nobreak\hfil
      \nobreak\hb@xt@\@pnumwidth{\hss #2}\par
    \endgroup
  \fi}
\renewcommand*\l@subsection{\@dottedtocline{2}{2.4em}{3.2em}}
\makeatother

\setcounter{topnumber}{4}
\setcounter{bottomnumber}{2}
\setcounter{totalnumber}{6}
\setcounter{dbltopnumber}{4}
\renewcommand{\topfraction}{0.92}
\renewcommand{\dbltopfraction}{0.92}
\renewcommand{\bottomfraction}{0.5}
\renewcommand{\textfraction}{0.06}
\renewcommand{\floatpagefraction}{0.75}
\renewcommand{\dblfloatpagefraction}{0.75}

\maketitlesupplementary
\addtocontents{toc}{\protect\setcounter{tocdepth}{3}}
\noindent This supplementary material accompanies the main paper. It
contains twelve sections (D1--D12) that the main paper cross-references
for more detailed low level information. Section
labels are referenced from the main paper as ``supplementary \S\,D$n$''.

\tableofcontents

\section{Vehicle blueprint to class mapping}
\label{sec:supp_vehicle_classes}

The 37-entry mapping from CARLA \texttt{vehicle.*} blueprints to the
hand-curated five-class taxonomy of
\texttt{vehicle\_class\_json\_file.txt} is reproduced verbatim in
Table~\ref{tab:vehicle_classes}. The taxonomy is
$\{0=\text{passenger car},\,1=\text{truck (incl.\ buses and fire
trucks)},\,2=\text{motorcycle},\,3=\text{bicycle},\,4=\text{bigger car
(vans, large SUVs, ambulances)}\}$. As discussed in the main paper, only classes 0, 1, and 4 are
populated in the released annotations: two-wheelers (classes 2, 3) were
deliberately excluded from the spawnable blueprint pool during
collection (the corresponding seven blueprints listed in
\texttt{excluded\_blueprints} of the collection script: Harley-Davidson
Low Rider, Yamaha YZF, Kawasaki Ninja, Vespa ZX-125, BH Crossbike,
Gazelle Omafiets, Diamondback Century). The gap is preserved so the
emitted IDs continue to match the hand-curated table.

\begin{table*}[t]
\centering
\footnotesize
\begin{tabular}{@{}lc|lc@{}}
\toprule
Blueprint id & Class & Blueprint id & Class \\
\midrule
\texttt{vehicle.audi.a2}                       & 0 & \texttt{vehicle.lincoln.mkz\_2020}             & 0 \\
\texttt{vehicle.nissan.micra}                  & 0 & \texttt{vehicle.citroen.c3}                    & 0 \\
\texttt{vehicle.audi.tt}                       & 0 & \texttt{vehicle.dodge.charger\_police}         & 0 \\
\texttt{vehicle.mercedes.coupe\_2020}          & 0 & \texttt{vehicle.nissan.patrol}                 & 4 \\
\texttt{vehicle.bmw.grandtourer}               & 0 & \texttt{vehicle.jeep.wrangler\_rubicon}        & 4 \\
\texttt{vehicle.harley-davidson.low\_rider}    & 2 & \texttt{vehicle.mini.cooper\_s}                & 0 \\
\texttt{vehicle.ford.ambulance}                & 4 & \texttt{vehicle.mercedes.coupe}                & 0 \\
\texttt{vehicle.micro.microlino}               & 0 & \texttt{vehicle.dodge.charger\_2020}           & 0 \\
\texttt{vehicle.carlamotors.firetruck}         & 1 & \texttt{vehicle.ford.crown}                    & 0 \\
\texttt{vehicle.carlamotors.carlacola}         & 1 & \texttt{vehicle.seat.leon}                     & 0 \\
\texttt{vehicle.ford.mustang}                  & 0 & \texttt{vehicle.toyota.prius}                  & 0 \\
\texttt{vehicle.chevrolet.impala}              & 0 & \texttt{vehicle.yamaha.yzf}                    & 2 \\
\texttt{vehicle.kawasaki.ninja}                & 2 & \texttt{vehicle.mitsubishi.fusorosa}           & 1 \\
\texttt{vehicle.tesla.model3}                  & 0 & \texttt{vehicle.tesla.cybertruck}              & 1 \\
\texttt{vehicle.mercedes.sprinter}             & 4 & \texttt{vehicle.audi.etron}                    & 0 \\
\texttt{vehicle.volkswagen.t2}                 & 4 & \texttt{vehicle.lincoln.mkz\_2017}             & 0 \\
\texttt{vehicle.dodge.charger\_police\_2020}   & 0 & \texttt{vehicle.vespa.zx125}                   & 2 \\
\texttt{vehicle.mini.cooper\_s\_2021}          & 0 & \texttt{vehicle.nissan.patrol\_2021}           & 0 \\
\texttt{vehicle.volkswagen.t2\_2021}           & 4 & & \\
\bottomrule
\end{tabular}
\caption{Complete CARLA \texttt{vehicle.*} blueprint to hand-curated
five-class taxonomy mapping (37 entries). Classes~2 and~3 are present
in the mapping but never instantiated during collection, so the released
annotations contain only classes~0, 1, and~4. Source:
\texttt{vehicle\_class\_json\_file.txt} (read at collection time via the
\texttt{JSON\_PATH} constant of the collection script).}
\label{tab:vehicle_classes}
\end{table*}

\section{Full directory tree}
\label{sec:supp_tree}

The complete on-disk directory tree of the released dataset is reproduced
below; the main paper refers to it from \S\,3.4 (``Accessing the
dataset'') rather than reproducing it. Each
\texttt{data\_collection\_<town>\_<tod>\_run<i>\_<id>/} folder holds one
contiguous collection run; within it, each camera directory holds the
clean RGB, three foggy variants, depth, segmentation, and 2D bbox
sub-directories; LiDAR and its three foggy variants, semantic LiDAR,
radar, 3D bboxes, and the per-scene extrinsic configuration live at the
collection-run level.

{\footnotesize
\begin{verbatim}
dataset_root/
|-- data_root/
|   |-- data_collection_T1_day_run1_<id>/
|   |   |-- cam_front/
|   |   |   |-- out_rgb/
|   |   |   |-- out_fog_light/
|   |   |   |-- out_fog_moderate/
|   |   |   |-- out_fog_dense/
|   |   |   |-- out_depth/
|   |   |   |-- out_seg/
|   |   |   `-- out_bbox/
|   |   |-- cam_back/    (identical sub-tree)
|   |   |-- cam_left/    (identical sub-tree)
|   |   |-- cam_right/   (identical sub-tree)
|   |   |-- lidar/
|   |   |-- lidar_fog_light/
|   |   |-- lidar_fog_moderate/
|   |   |-- lidar_fog_dense/
|   |   |-- semantic_lidar/
|   |   |-- radar/
|   |   |-- out_3dbbox/
|   |   `-- extrinsic_config/
|   |-- data_collection_T1_night_run1_<id>/
|   |-- ...
|   `-- data_collection_T10_night_run<k>_<id>/
`-- index/
    |-- scene.json
    |-- sample.json
    |-- sample_data.json
    `-- splits.json
\end{verbatim}}

Filename fields in \texttt{sample\_data.json} are relative paths starting
with \texttt{data\_root/} and are resolved as \texttt{dataroot / filename}
by the access class.

\section{3D-to-2D bounding-box projection: full derivation}
\label{sec:supp_projection}

2D bounding boxes are obtained by projecting the released 3D boxes into each
camera. 3D boxes are stored in the LiDAR sensor frame (CARLA convention:
$+X$ forward, $+Y$ right, $+Z$ up, in metres); 2D boxes are pixel-space
axis-aligned rectangles in the computer-vision convention ($+X$ right,
$+Y$ down, $+Z$ into the image). The projection composes a per-scene
extrinsic $E$ with a single shared intrinsic
\begin{align}
K = \begin{bmatrix} f_x & 0 & c_x \\ 0 & f_y & c_y \\ 0 & 0 & 1 \end{bmatrix}
  = \begin{bmatrix} 537.02 & 0 & 640 \\ 0 & 537.02 & 480 \\ 0 & 0 & 1 \end{bmatrix},
\label{eq:supp_intrinsic}
\end{align}
where $f_x = f_y = (W/2)/\tan(\mathrm{FOV}/2) \approx 537.02$ and
$(c_x, c_y) = (640, 480)$ for a $W \times H = 1280\times960$ image at
$\mathrm{FOV} = 100^{\circ}$. The full projection is then:

\begin{align}
P_{\mathrm{lidar}}      &= [x, y, z, 1]^{\top},                                \label{eq:lidar} \\
P_{\mathrm{cam}}        &= E \cdot P_{\mathrm{lidar}}, \notag \\
P_{\mathrm{unnorm}}     &= K \cdot [x_{\mathrm{cam}}, y_{\mathrm{cam}}, z_{\mathrm{cam}}]^{\top}
                          = [u', v', w']^{\top},                              \label{eq:unnorm} \\
u                       &= u'/w', \quad v = v'/w',                            \notag \\
P_{\mathrm{pixel}}      &= [\,\mathrm{round}(u),\, \mathrm{round}(v)\,]^{\top}.\notag
\end{align}

Applying this to all eight corners $\{(u_i, v_i)\}_{i=1}^{8}$ of a 3D box and
taking their axis-aligned hull gives the released 2D box,
$[[\min_i u_i,\, \min_i v_i],\, [\max_i u_i,\, \max_i v_i]]$, clipped to the
image bounds.

The intrinsic depends only on camera configuration, which is identical
across the dataset, so $K$ is shared by every camera and every scene.
The extrinsic depends on the relative pose of camera and LiDAR; the
Unreal Engine spawn process introduces small per-scene jitter, so $E$ is
released per scene (one matrix per camera-LiDAR pair, in the per-scene
\texttt{extrinsic\_config.json}).

Note that the released $E$ maps LiDAR coordinates \emph{directly} into the
vision-convention camera frame: it composes the rigid LiDAR-to-camera
transform with the CARLA-to-vision axis permutation
$(x, y, z) \mapsto (y, -z, x)$, which is why
Eq.~\eqref{eq:unnorm} applies $K$ to $P_{\mathrm{cam}}$ with no further
reordering. Users who rebuild $E$ themselves from the sensor poses of
\S\ref{sec:supp_sensor} must include that permutation.

\begin{figure}[t]
  \centering
  \includegraphics[width=0.92\linewidth]{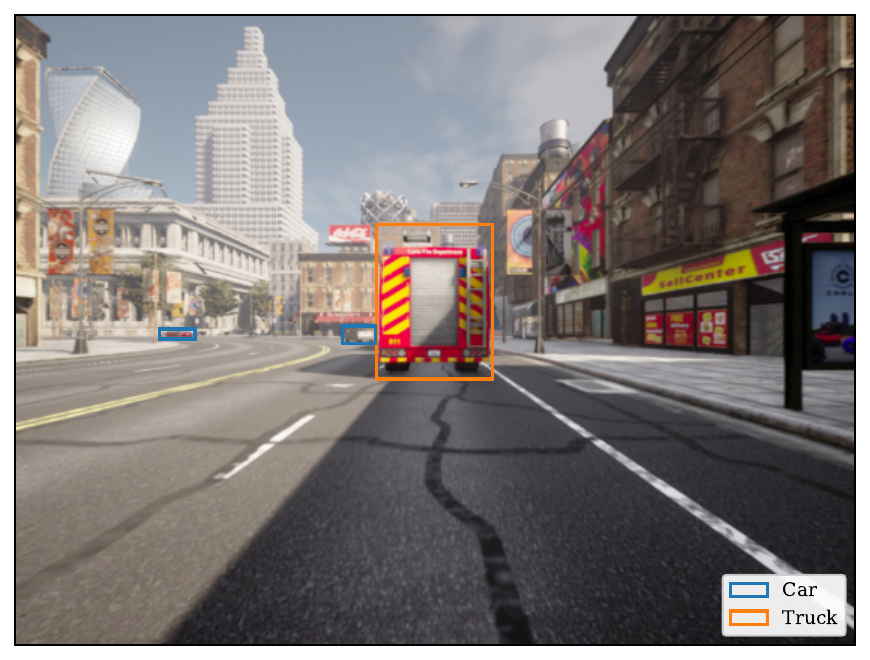}
  \caption{Released 2D boxes for one front-camera frame, coloured by class.
  Boxes are the axis-aligned hulls of the eight projected 3D-box corners.}
  \label{fig:supp_bbox_overlay}
\end{figure}

An example of the resulting 2D boxes is shown in
Figure~\ref{fig:supp_bbox_overlay}.

\paragraph{Multi-camera visibility extension.}
The open-source CARLA-2DBBox annotator originally supported a single camera
placed identically to the LiDAR for visibility checks. We extended it so that
all four cameras of differing orientation share one semantic LiDAR for
visibility determination, keeping annotation a single sweep over the scene
rather than four. The \texttt{min\_detect}~$=5$-point and
\texttt{max\_dist}~$=\SI{100}{m}$ criteria are identical for 2D and 3D boxes.
The limitation is geometric: the cameras must remain close to the LiDAR,
since a vehicle visible to a camera but occluded from the LiDAR (or vice
versa) will be mis-annotated; the main paper's annotation-quality section
quantifies this effect.

\paragraph{Note on near-plane clipping.}
The reference projection routine does not clip the 3D bounding box
against the camera near plane before computing the axis-aligned 2D box.
When a vehicle passes very close to the ego camera, one or more 3D-box
corners can have non-positive $z_{\mathrm{cam}}$ and the perspective
divide above produces a 2D AABB that spans most of the image. We
quantify the prevalence ($\approx \SI{0.8}{\percent}$ of annotated
boxes) in the main paper's annotation-quality section, and we provide a
recommended read-time heuristic for downstream users:

\begin{quote}
\textit{Drop any 2D bounding box whose area is $\geq \SI{50}{\percent}$
of the image area \emph{or} that touches at least three image edges.}
\end{quote}

\section{Access class: complete method signatures}
\label{sec:supp_access}

The main paper gives a high-level overview of the \texttt{FogDrive}
access class; the complete user-facing method list is reproduced here.
The class loads the four metadata JSONs (\texttt{scene.json},
\texttt{sample.json}, \texttt{sample\_data.json}, \texttt{splits.json})
at construction time and builds the two pre-computed joins described in
the main paper.

\paragraph{O(1) lookup methods (constant-time once the index is built).}
\begin{itemize}
\item \texttt{get(table\_name, token)} --- returns the record dict for a
  given token in \texttt{scene} / \texttt{sample} / \texttt{sample\_data}.
\item \texttt{get\_idx(table\_name, token)} --- returns the integer
  index of that token in its table list.
\item \texttt{get\_scene\_by\_name(name)} --- returns the scene record for a
  human-readable name (e.g.\ \texttt{T1\_day\_scene001}).
\item \texttt{get\_sample\_data\_path(sample\_data\_token)} --- returns
  the absolute path to the file for a sample\_data token.
\item \texttt{get\_extrinsic\_config\_path(scene\_token)} --- returns
  the absolute path to the scene's extrinsic-config JSON;
  \texttt{load\_extrinsic\_config(scene\_token)} additionally reads and
  parses it into a Python dict.
\end{itemize}

\paragraph{Linear-in-output methods (dict lookups + pre-computed lists; no scan of raw tables).}
\begin{itemize}
\item \texttt{get\_sample\_data\_for\_sample(sample\_token, channels=None)}
  --- full sample\_data records for one sample, optionally filtered to a
  channel list.
\item \texttt{get\_sample\_paths(sample\_token, channels=None,
  absolute=True)} --- the same as a channel-to-path dict (the fast path
  for training loaders).
\item \texttt{get\_samples\_for\_scene(scene\_token)} --- all sample
  records in chronological order.
\item \texttt{get\_scenes\_for\_split(split)},
  \texttt{get\_samples\_for\_split(split)},
  \texttt{get\_sample\_tokens\_for\_split(split)} --- scenes / samples /
  sample-tokens for \texttt{train} / \texttt{val} / \texttt{test}
  (requires \texttt{splits.json}).
\item \texttt{list\_channels()}, \texttt{list\_scene\_names()} --- sorted
  list of unique channel names / scene names.
\end{itemize}

\paragraph{Linear-in-table fallback (intended for ad-hoc inspection, not training loops).}
\begin{itemize}
\item \texttt{field2token(table\_name, field, query)} --- scans the
  named table for records whose field equals the query value and
  returns their tokens. O(table size).
\end{itemize}

A reference \texttt{FogDriveDataset} PyTorch wrapper supporting channel
filtering and split selection is also provided.

\paragraph{Index cost.}
The constant-time guarantees above are paid for once at construction, when
the four metadata tables are parsed and the reverse indices over all
\NumSampleData{} \texttt{sample\_data} records are built. The index is
metadata only: image and point-cloud payloads stay on disk until a path
returned by the access class is explicitly read.

\section{Verbose sensor attribute tables}
\label{sec:supp_sensor}

The main paper describes the sensor rig in prose and shows its layout in
bird's-eye view. Table~\ref{tab:sensor_pose} gives the numerical poses of
every sensor, and Table~\ref{tab:supp_sensor_attr} the full per-sensor
attribute list, for users who wish to instantiate a matching rig in their
own CARLA pipeline. All values are read from the collection script
(\texttt{collect\_and\_annotate\_data\_1.py}, sensor-blueprint
configuration block).

\begin{table*}[t]
\centering
\footnotesize
\setlength{\tabcolsep}{4pt}
\begin{tabular}{@{}lccl@{}}
\toprule
Sensor & $(x,y,z)$~[m] & (pitch, roll, yaw)~[deg] & Key non-default attributes \\
\midrule
Front cam (RGB/depth/seg)  & $(1.8,~0,~1.2)$    & $(0,~0,~0)$    & $1280\times960$\,px, H-FOV~$100^{\circ}$ \\
Right cam (RGB/depth/seg)  & $(0.5,~0.7,~1.5)$  & $(0,~0,~100)$  & $1280\times960$\,px, H-FOV~$100^{\circ}$ \\
Left cam (RGB/depth/seg)   & $(0.5,~-0.7,~1.5)$ & $(0,~0,~-100)$ & $1280\times960$\,px, H-FOV~$100^{\circ}$ \\
Back cam (RGB/depth/seg)   & $(-2.4,~0,~1.2)$   & $(0,~0,~180)$  & $1280\times960$\,px, H-FOV~$100^{\circ}$ \\
LiDAR                      & $(0,~0,~2)$        & $(0,~0,~0)$    & 64 ch, \SI{120}{m}, 1.3\,M pts/s, \SI{20}{Hz}, V-FOV~$[-20,+5]^{\circ}$ \\
Semantic LiDAR             & $(0,~0,~2)$        & $(0,~0,~0)$    & 64 ch, \SI{120}{m}, 750k pts/s, \SI{20}{Hz}, V-FOV~$[-20,+5]^{\circ}$ \\
Radar                      & $(2,~0,~1.2)$      & $(0,~0,~0)$    & \SI{150}{m}, 18k pts/s, H/V-FOV~$120^{\circ}/30^{\circ}$ \\
\bottomrule
\end{tabular}
\caption{Sensor poses in the ego frame (CARLA's left-handed $+X$-forward,
$+Y$-right, $+Z$-up convention) with key non-default attributes. The depth
and semantic-segmentation cameras share pose and attributes with the
co-located RGB camera. Referenced from the main paper's sensor-configuration
section.}
\label{tab:sensor_pose}
\end{table*}

\begin{table*}[t]
\centering
\footnotesize
\begin{tabular}{@{}lll@{}}
\toprule
Sensor & Attribute & Value \\
\midrule
\multirow{8}{*}{LiDAR (\texttt{sensor.lidar.ray\_cast})}
 & Channels                    & 64 \\
 & Range                       & \SI{120}{m} \\
 & Points per second           & 1\,300\,000 \\
 & Rotation frequency          & \SI{20}{Hz} \\
 & Upper FOV                   & $+5^{\circ}$ \\
 & Lower FOV                   & $-20^{\circ}$ \\
 & Horizontal FOV              & $360^{\circ}$ \\
 & Other attributes            & blueprint defaults \\
\midrule
\multirow{8}{*}{Semantic LiDAR (\texttt{sensor.lidar.ray\_cast\_semantic})}
 & Channels                    & 64 \\
 & Range                       & \SI{120}{m} \\
 & Points per second           & 750\,000 \\
 & Rotation frequency          & \SI{20}{Hz} \\
 & Upper FOV                   & $+5^{\circ}$ \\
 & Lower FOV                   & $-20^{\circ}$ \\
 & Horizontal FOV              & $360^{\circ}$ \\
 & Other attributes            & blueprint defaults \\
\midrule
\multirow{5}{*}{Radar (\texttt{sensor.other.radar})}
 & Range                       & \SI{150}{m} \\
 & Points per second           & 18\,000 \\
 & Horizontal FOV              & $120^{\circ}$ \\
 & Vertical FOV                & $30^{\circ}$ \\
 & Other attributes            & blueprint defaults \\
\midrule
\multirow{3}{*}{RGB camera (\texttt{sensor.camera.rgb})}
 & Image size $(x, y)$         & $(1280, 960)$\,px \\
 & Horizontal FOV              & $100^{\circ}$ \\
 & Other attributes            & blueprint defaults \\
\midrule
\multirow{3}{*}{Depth camera (\texttt{sensor.camera.depth})}
 & Image size $(x, y)$         & $(1280, 960)$\,px \\
 & Horizontal FOV              & $100^{\circ}$ \\
 & Other attributes            & blueprint defaults \\
\midrule
\multirow{3}{*}{Semantic-seg camera (\texttt{sensor.camera.semantic\_segmentation})}
 & Image size $(x, y)$         & $(1280, 960)$\,px \\
 & Horizontal FOV              & $100^{\circ}$ \\
 & Other attributes            & blueprint defaults \\
\bottomrule
\end{tabular}
\caption{Full per-sensor attribute table. The depth and
semantic-segmentation cameras share image size and FOV with the RGB
camera (and the same pose), so they appear under three rows each.}
\label{tab:supp_sensor_attr}
\end{table*}

\section{Verbose metadata JSON schemas}
\label{sec:supp_json}

The main paper gives a compact summary of the four metadata JSON tables.
The full per-field schema is reproduced below.

\paragraph{\texttt{scene.json}.} Top-level list; one dict per scene.
\begin{itemize}
\item \texttt{token} --- \texttt{str}; uuid4, unique to this scene.
\item \texttt{name} --- \texttt{str}; human-readable scene name, format
  \texttt{T<town>\_<day|night>\_scene<NNN>}.
\item \texttt{description} --- \texttt{str}; comma-separated tags for
  town number, time of day, and run number.
\item \texttt{extrinsic\_config\_path} --- \texttt{str}; path (relative
  to dataroot) to the per-scene extrinsic-matrix JSON.
\item \texttt{nbr\_samples} --- \texttt{int}; number of samples in the
  scene (\SI{0.1}{s} apart, so seconds-per-scene $= 0.1 \times$ this).
\item \texttt{first\_sample\_token} --- \texttt{str}; token of the first
  sample of this scene.
\item \texttt{last\_sample\_token} --- \texttt{str}; token of the last
  sample of this scene.
\end{itemize}

\paragraph{\texttt{sample.json}.} Top-level list; one dict per sample
(frame).
\begin{itemize}
\item \texttt{token} --- \texttt{str}; uuid4, unique to this sample.
\item \texttt{timestamp} --- \texttt{float}; microseconds since
  scene start.
\item \texttt{scene\_token} --- \texttt{str}; token of the scene this
  sample belongs to.
\item \texttt{prev} --- \texttt{str}; token of the previous sample in
  the scene (empty string at first).
\item \texttt{next} --- \texttt{str}; token of the next sample in the
  scene (empty string at last).
\end{itemize}

\paragraph{\texttt{sample\_data.json}.} Top-level list; one dict per
sensor reading (sample $\times$ channel).
\begin{itemize}
\item \texttt{token} --- \texttt{str}; uuid4, unique to this reading.
\item \texttt{sample\_token} --- \texttt{str}; token of the sample this
  reading belongs to.
\item \texttt{channel} --- \texttt{str}; the sensor-channel identifier
  (e.g.\ \texttt{CAM\_FRONT}, \texttt{CAM\_FRONT\_FOG\_DENSE},
  \texttt{LIDAR\_TOP}, \texttt{LIDAR\_TOP\_FOG\_LIGHT},
  \texttt{SEM\_LIDAR\_TOP}, \texttt{RADAR\_FRONT},
  \texttt{CAM\_FRONT\_BBOX}, \texttt{CAM\_FRONT\_DEPTH},
  \texttt{CAM\_FRONT\_SEM}, \texttt{BBOX\_3D}); full list in the main
  paper's channels table.
\item \texttt{filename} --- \texttt{str}; path to the data file,
  relative to the dataroot.
\item \texttt{fileformat} --- \texttt{str}; one of \texttt{png},
  \texttt{npy}, \texttt{json}.
\item \texttt{width}  --- \texttt{int}; PNG only, width of the image
  in pixels.
\item \texttt{height} --- \texttt{int}; PNG only, height of the image
  in pixels.
\end{itemize}

\paragraph{\texttt{splits.json}.} Top-level dict with keys
\texttt{train}, \texttt{val}, \texttt{test}; each value is a list of
human-readable scene names. The dataset is split 70/15/15 with the
day/night ratio held exactly at 50/50 within every split, and per-town
shares matched to the overall distribution within $\pm 2$ percentage
points. Split coverage checks (\texttt{duplicate\_assignments},
\texttt{scenes\_in\_no\_split},
\texttt{scenes\_in\_split\_but\_not\_in\_index}) all pass empty: every
scene in the index is assigned to exactly one split, and every split
references only indexed scenes. Table~\ref{tab:splits} gives the full
split sizes.

\begin{table}[t]
\centering
\scriptsize
\setlength{\tabcolsep}{3pt}
\begin{tabular}{@{}lrrrcc@{}}
\toprule
Split & Scenes & Frames & Scene fraction & Day/Night & T1/T2/T3/T4/T5/T10 share \\
\midrule
train & 466 & 93\,058 & 70.0\,\% & 233 / 233 & matched to overall \\
val   & 100 & 19\,968 & 15.0\,\% &  50 /  50 & matched to overall \\
test  & 100 & 19\,970 & 15.0\,\% &  50 /  50 & matched to overall \\
\midrule
\textbf{total} & \textbf{666} & \textbf{132\,996} & \textbf{100\,\%} & \textbf{333 / 333} & 14/7/21/24/24/10\,\% \\
\bottomrule
\end{tabular}
\caption{Train/val/test split sizes; day/night held 50/50 and per-town
shares matched across splits.}
\label{tab:splits}
\end{table}

\section{Per-modality disk-footprint breakdown}
\label{sec:supp_disk}

The main paper reports the aggregate size of FogDrive
($\approx 2.74$\,TiB) and the camera-channel share ($\approx 86\,\%$).
The full per-modality breakdown is given in
Table~\ref{tab:supp_disk_breakdown}: file counts and on-disk sizes for
each sensor / fog-level combination. Annotation cost (2D + 3D
bounding-box JSON combined) is negligible at $<1$\,GB.

\begin{table*}[t]
\centering
\footnotesize
\begin{tabular}{@{}lrr@{}}
\toprule
Modality & Files & Size (GB) \\
\midrule
Camera RGB (4 cameras, clean)            & 531\,984 & 611.3 \\
Camera RGB (light / moderate / dense fog) & $3 \times 531\,984$ & 500.0 / 479.1 / 430.4 \\
Camera depth                              & 531\,984 & 387.4 \\
Camera segmentation                       & 531\,984 &  11.9 \\
LiDAR (clean)                             & 132\,996 &  67.0 \\
LiDAR (light / moderate / dense fog)      & $3 \times 132\,996$ & 68.0 / 70.6 / 76.3 \\
Semantic LiDAR                            & 132\,996 & 105.6 \\
Radar                                     & 132\,996 &   1.6 \\
2D bbox annotations                       & 531\,984 &   0.2 \\
3D bbox annotations                       & 132\,996 &   0.9 \\
\midrule
\textbf{Total}                            & \textbf{4\,654\,860} & \textbf{2\,810.3} \\
\bottomrule
\end{tabular}
\caption{On-disk footprint per modality at the full dataset scale.}
\label{tab:supp_disk_breakdown}
\end{table*}

\section{Per-sensor file formats}
\label{sec:supp_formats}

The complete per-modality on-disk format specification --- array
shapes, channel conventions, the depth-encoding equation, and the
2D / 3D bounding-box JSON schemas --- is given here; the main paper
refers to this section rather than summarising it.

\paragraph{Camera channels.}
Each of the four cameras produces, per frame, a clean RGB PNG; three
foggy RGB PNG variants (light, moderate, dense); a depth PNG with
distance encoded as
\begin{align*}
\mathrm{normalised} &= (R + 256\,G + 256^{2}\,B)\,/\,(256^{3} - 1), \\
\mathrm{metres}     &= 1000 \times \mathrm{normalised},
\end{align*}
which uses the full 24-bit-per-pixel space and gives sub-millimetre
depth precision (relevant in fog because depth errors propagate into
image transmission via $t = \exp(-\beta\,d)$); a semantic-segmentation
PNG (greyscale, red channel = CARLA class id); and a 2D bounding-box
JSON.

\paragraph{2D bounding-box JSON.}
Contains \texttt{bboxes} --- an $(N, 2, 2)$ integer list of
$[[x_{\min}, y_{\min}], [x_{\max}, y_{\max}]]$ pairs in pixel
coordinates; \texttt{vehicle\_ids} --- per-scenario unique id per box;
and \texttt{vehicle\_class} --- the taxonomy id of each detected
vehicle.

\begin{figure}[t]
  \centering
  \includegraphics[width=\linewidth]{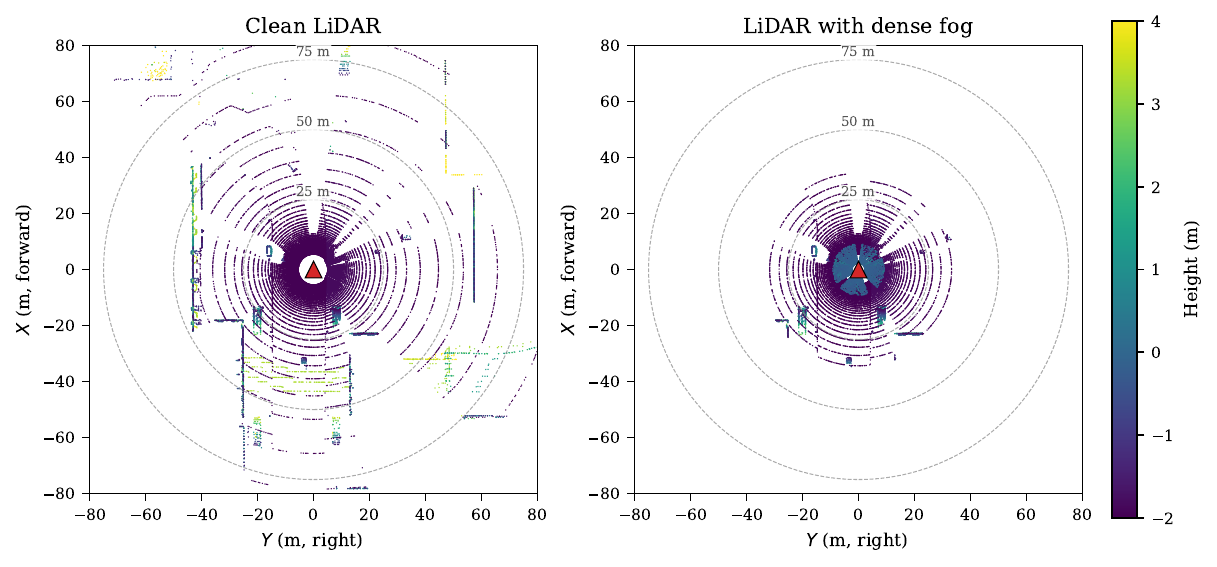}
  \caption{A released LiDAR sweep under dense fog, bird's-eye view. Fog
  backscatter concentrates returns near the sensor and truncates the usable
  range.}
  \label{fig:supp_lidar_bev}
\end{figure}

\begin{figure*}[t]
  \centering
  \includegraphics[width=\linewidth]{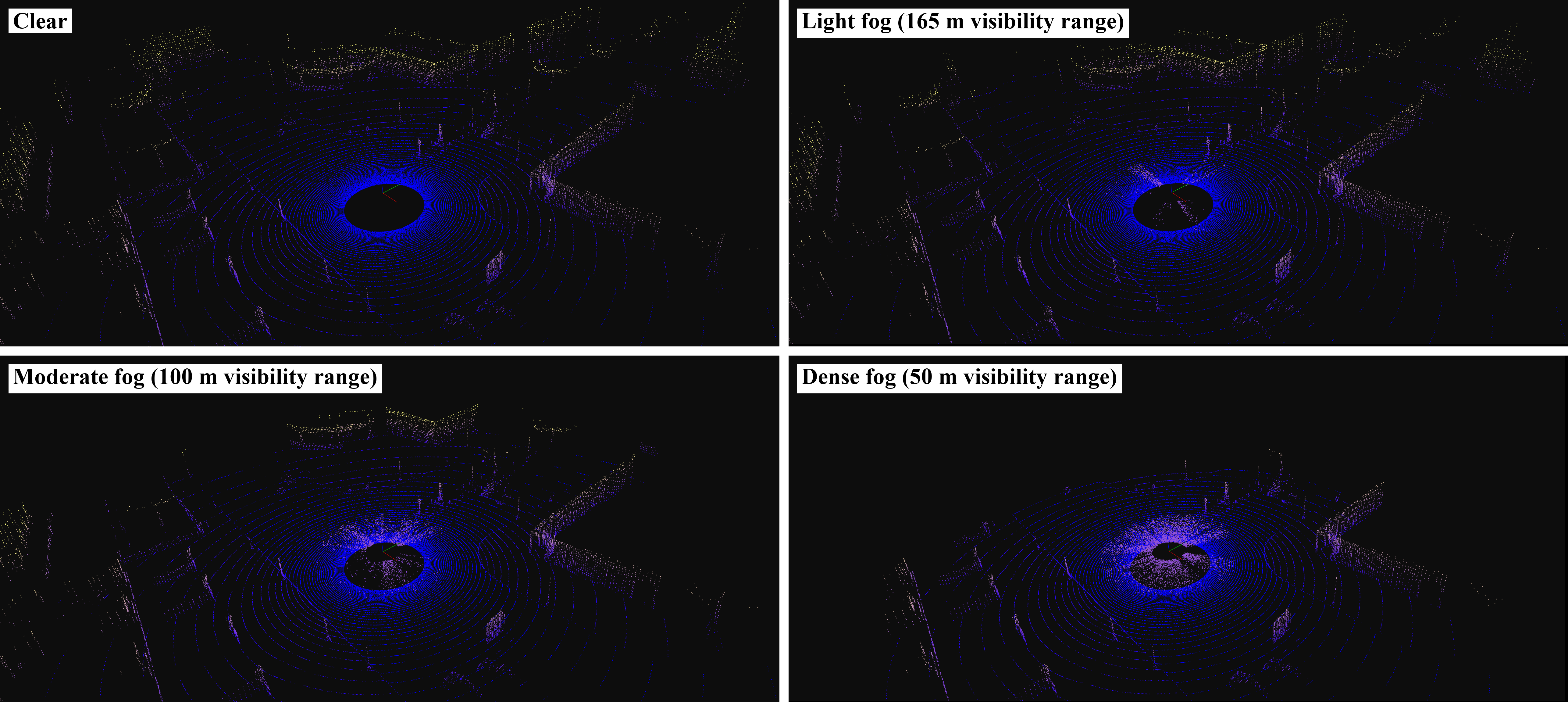}
  \caption{One released LiDAR sweep at all four densities, rendered from a
  fixed viewpoint and coloured by height. As density rises the returns
  concentrate into near-sensor backscatter and the far field thins out.}
  \label{fig:supp_lidar_levels}
\end{figure*}

\paragraph{LiDAR / semantic LiDAR / radar.}
LiDAR (clean and the three fog variants) is stored as a
\texttt{float32} $(N, 4)$ \texttt{.npy} array
$(x, y, z, \mathrm{intensity})$. Semantic LiDAR is stored as
$(N, 6)$ $(x, y, z, \cos\theta, \mathrm{id}, \mathrm{tag})$ with the
\texttt{uint32} CARLA \texttt{id} and \texttt{tag} fields cast to
\texttt{float32} before saving. Radar is stored as $(N, 4)$
$(\mathit{velocity}, \mathit{azimuth}, \mathit{altitude},
\mathit{depth})$ from CARLA's \texttt{RadarMeasurement} buffer, where
\emph{depth} $d$ is the radial range in metres and the angles are in
radians from the sensor's forward axis. Cartesian coordinates in the
sensor frame are recovered as $x = d\cos\phi\cos\theta$,
$y = d\cos\phi\sin\theta$, $z = d\sin\phi$ with $\theta = \mathit{azimuth}$
and $\phi = \mathit{altitude}$. Figure~\ref{fig:supp_lidar_bev} shows a
released LiDAR sweep under dense fog in bird's-eye view.
Figure~\ref{fig:supp_lidar_levels} shows the same sweep at all four released
densities from a fixed viewpoint. Table~\ref{tab:radar_fog} collects the
published measurements behind the main paper's decision to release radar
without fog degradation.

\begin{table}[t]
\centering
\scriptsize
\setlength{\tabcolsep}{2.6pt}
\begin{tabular}{@{}llll@{}}
\toprule
Sensor & Condition & Effect on signal & Source \\
\midrule
Radar \SI{77}{GHz} & dense fog       & $<\SI{1}{dB/km}$                              & ITU-R P.840 \\
Radar \SI{77}{GHz} & mod./dense fog  & ``largely unaffected''                        & Srivastav '23 \\
Radar \SI{77}{GHz} & fog backscatter & negligible                                    & Bijelic '20 \\
LiDAR \SI{905}{nm} & dense fog       & usable range $<\SI{70}{m}$                     & El Haddioui '25 \\
\bottomrule
\end{tabular}
\caption{Evidence that fog affects \SI{77}{GHz} radar far less than LiDAR,
motivating the undegraded radar release; appreciable fog attenuation sets in
only near \SI{100}{GHz}. Full citations accompany this claim in the main
paper's degradation section.}
\label{tab:radar_fog}
\end{table}

\paragraph{3D bounding-box JSON.}
Released per frame as a JSON file containing \texttt{bboxes\_3d} ---
an $(N, 8, 3)$ float list giving all eight corner points of each
vehicle's box in metres in the LiDAR sensor frame (storing all eight
preserves vehicle orientation rather than just an axis-aligned extent)
--- and \texttt{vehicle\_ids}.

\paragraph{Per-scene extrinsic configuration.}
A per-scene \texttt{extrinsic\_config.json} stores the four
LiDAR-to-camera $4\times 4$ extrinsic matrices under the keys
\texttt{extrinsic\_\{front, right, left, back\}}; their use in the
3D-to-2D projection is given in the main paper's
\S\,3.3 and in supplementary~\S\,D3.

\section{Scenario-collection implementation notes}
\label{sec:supp_collection_notes}

The main paper notes that FogDrive scenarios are distributed across six
CARLA towns. Two collection-time details that affect what ends up in
the released annotations are recorded here for reproducibility.

\paragraph{Parked vehicles.}
Every map is loaded in its \texttt{\_Opt} (``optional layers'') variant
so that the \texttt{ParkedVehicles} map layer can be unloaded at the
start of every scenario. CARLA exposes parked vehicles as static map
geometry rather than as \texttt{vehicle.*} actors, which places them
outside the visibility test used by the annotator
(\S\,\ref{sec:supp_projection}); unloading them guarantees that
every car visible in the released images is also represented in the
annotations.

\paragraph{Traffic-light phasing.}
Traffic-light cycles are overridden to a 5\,/\,\SI{2.5}{s}
green/yellow schedule so that the ego does not idle at a red light for
the full 20-second scenario window.

\paragraph{Background-traffic density.}
Scenarios are captured in CARLA's synchronous mode with
\texttt{fixed\_delta\_seconds}~$=0.05$ (\SI{20}{Hz} simulation, keyframes
saved at \SI{10}{Hz}). Background traffic is spawned up to a town-dependent
NPC cap, scaled to map size: \texttt{30} for Town02, \texttt{40} for
Town10HD, \texttt{60} for Town01, \texttt{90} for Town03, and \texttt{100}
for Town04 and Town05.
All NPCs are drawn from the full CARLA \texttt{vehicle.*} blueprint pool
minus the seven excluded two-wheelers and handed to the same Traffic
Manager instance that drives the ego.

\paragraph{Time-of-day implementation.}
``Day'' uses CARLA's default weather preset; ``night'' is realised by
overriding the world's \texttt{sun\_altitude\_angle} to $-90^{\circ}$,
which captures the principal change in ambient illumination.

\section{Experiment 1: additional 3D-detection metrics}
\label{sec:supp_exp1}

The main paper (\S 4.1) reports mAP@0.5 and Mean TP for both detectors.
Table~\ref{tab:supp_exp1_ate} gives the
corresponding ATE ($1 - \text{translation error}/\SI{2}{m}$) matrices; the
scale (ASE) and orientation (AOE) true-positive components exhibit the same
Mixed-L advantage at smaller magnitude and follow the ATE pattern.

\paragraph{Training details.}
Both detectors share the VoxelResBackBone8x trunk and TransFusion detection
head; BEVFusion additionally fuses a Swin-T image backbone through a
depth-LSS view transformer.
TransFusion-L is trained for 50 epochs (batch size 32, AdamW one-cycle, peak
LR~$10^{-3}$) and BEVFusion for 15 epochs (batch size 4--8, peak
LR~$10^{-4}$). The point-cloud range is $\pm\SI{40}{m}$ for both. For every
(training-subset $\times$ test-subset) pair we report the checkpoint that
maximises the composite FogDrive Detection Score
($\mathrm{FDS} = 0.5\,\mathrm{mAP} + 0.5\,\mathrm{Mean\,TP}$) on that test
subset across the training schedule.

\begin{table}[t]
\centering
\footnotesize
\setlength{\tabcolsep}{4pt}
\begin{tabular}{@{}lccc|ccc@{}}
\toprule
 & \multicolumn{3}{c|}{TransFusion-L} & \multicolumn{3}{c}{BEVFusion} \\
Train$\downarrow$ / Test$\rightarrow$ & Light & Mod. & Dense & Light & Mod. & Dense \\
\midrule
Mixed-L  & \textbf{0.9473} & \textbf{0.9457} & \textbf{0.9474} & \textbf{0.9536} & \textbf{0.9441} & \textbf{0.9513} \\
Light    & 0.9446 & 0.9265 & 0.9425 & 0.9491 & 0.9344 & 0.9442 \\
Mixed-S  & 0.9438 & 0.9395 & 0.9451 & 0.9401 & 0.9323 & 0.9406 \\
Dense    & 0.9393 & 0.9346 & 0.9441 & 0.9220 & 0.9203 & 0.9313 \\
\bottomrule
\end{tabular}
\caption{Experiment~1, \textbf{ATE} ($1 - \text{translation error}/\SI{2}{m}$;
higher is better; column-wise winner in \textbf{bold}).}
\label{tab:supp_exp1_ate}
\end{table}

\begin{table*}[!t]
\centering
\scriptsize
\setlength{\tabcolsep}{6pt}
\begin{tabular}{@{}llrrrrrrr@{}}
\toprule
Pipeline & Fog & mAP@0.5 & mAP@.5:.95 & P & R & car & truck & l.veh \\
\midrule
A (clean$\to$clean) & clean & 0.692 & 0.539 & 0.808 & 0.623 & 0.839 & 0.498 & 0.740 \\
\midrule
B (clean$\to$fog) & light & 0.561 & 0.438 & 0.818 & 0.558 & 0.779 & 0.332 & 0.572 \\
B & moderate & 0.459 & 0.359 & 0.775 & 0.440 & 0.700 & 0.244 & 0.434 \\
B & dense & 0.272 & 0.215 & 0.689 & 0.253 & 0.440 & 0.092 & 0.284 \\
\midrule
C -- DCP & light & 0.627 & 0.490 & 0.794 & 0.606 & 0.826 & 0.368 & 0.687 \\
C -- DCP & moderate & 0.557 & 0.435 & \textbf{0.812} & 0.555 & 0.777 & 0.337 & 0.558 \\
C -- DCP & dense & 0.438 & 0.342 & 0.772 & 0.435 & 0.652 & 0.266 & 0.397 \\
C -- DehazeFormer-S & light & 0.585 & 0.461 & \textbf{0.831} & 0.571 & 0.803 & 0.349 & 0.601 \\
C -- DehazeFormer-S & moderate & 0.501 & 0.394 & 0.796 & 0.490 & 0.734 & 0.295 & 0.474 \\
C -- DehazeFormer-S & dense & 0.327 & 0.263 & 0.779 & 0.312 & 0.515 & 0.165 & 0.301 \\
C -- DehazeFormer-S (ft.) & light & 0.609 & 0.467 & 0.800 & 0.595 & 0.781 & 0.415 & 0.631 \\
C -- DehazeFormer-S (ft.) & moderate & 0.559 & 0.426 & 0.793 & 0.532 & 0.740 & 0.373 & 0.563 \\
C -- DehazeFormer-S (ft.) & dense & 0.415 & 0.328 & 0.771 & 0.408 & 0.598 & 0.267 & 0.380 \\
\midrule
\textbf{D (fog$\to$fog)} & light & \textbf{0.693} & \textbf{0.557} & 0.826 & \textbf{0.638} & \textbf{0.849} & \textbf{0.522} & \textbf{0.709} \\
\textbf{D} & moderate & \textbf{0.716} & \textbf{0.571} & 0.773 & \textbf{0.673} & \textbf{0.858} & \textbf{0.592} & \textbf{0.698} \\
\textbf{D} & dense & \textbf{0.647} & \textbf{0.510} & \textbf{0.851} & \textbf{0.591} & \textbf{0.867} & \textbf{0.400} & \textbf{0.674} \\
\bottomrule
\end{tabular}
\caption{Experiment~2 full detection metrics on the 7\,188-frame test split
per fog level. \textbf{Bold} marks the best value in each column among rows
comparable on the same test input (excluding A's clean row). Per-class
columns are AP for car, truck, and large vehicle.}
\label{tab:supp_exp2_full}
\end{table*}

\section{Annotation-audit details}
\label{sec:supp_audit}

\begin{figure*}[t]
  \centering
  \includegraphics[width=0.93\linewidth]{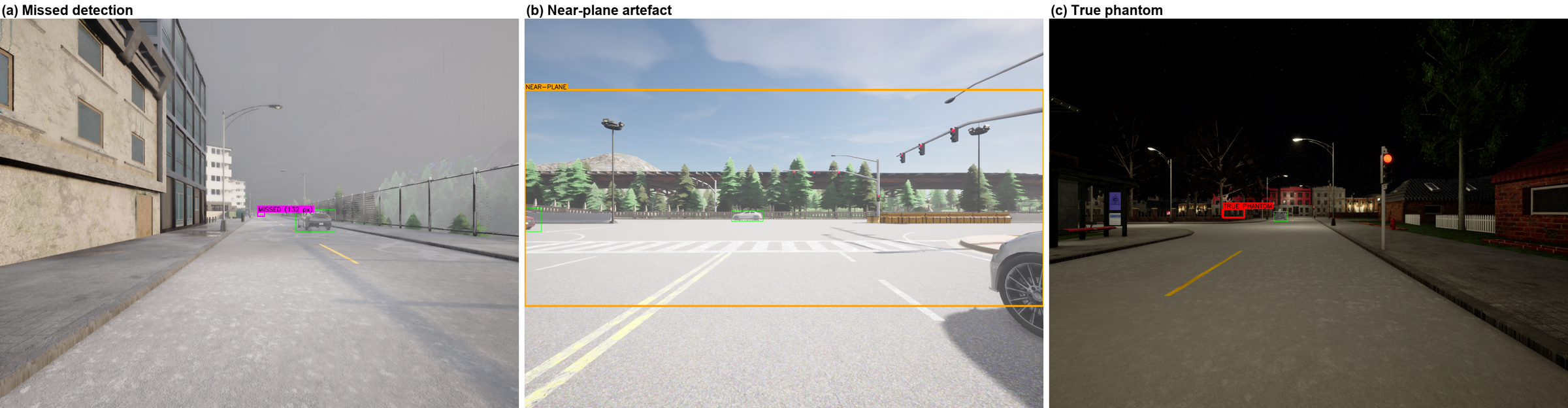}
  \caption{The three annotation failure modes, with released 2D boxes
  overlaid. (a)~A distant vehicle returning fewer than five semantic-LiDAR
  points is not annotated. (b)~A vehicle at the camera near plane yields an
  unclipped box spanning the frame. (c)~The LiDAR clips through an occluder
  the camera cannot see past, producing a box over no visible vehicle.}
  \label{fig:supp_failures}
\end{figure*}

This section gives the full audit protocol, threshold definitions,
distance-resolved recall, false-positive analysis and per-camera breakdown
behind the headline numbers in the main paper's annotation-quality section
(\S 3.7). Figure~\ref{fig:supp_failures} shows one representative example of
each of the three failure modes discussed below.

\paragraph{Protocol and thresholds.}
The audit compares the released LiDAR-derived 2D annotations against CARLA's
semantic-segmentation camera, which labels each pixel from scene metadata and
so shares no machinery with the visibility filter under test. For each of the
8\,096 audited images we build a binary vehicle-pixel mask and compute two
quantities against thresholds fixed \emph{before} any result was seen:
\begin{itemize}
\item \emph{Recall} --- for every connected vehicle blob of at least
  $A_{\min} = 100$\,px$^{2}$, whether some annotated box covers
  $\tau_{\mathrm{match}} = \SI{50}{\percent}$ or more of the blob's pixels.
  Blobs within 8\,px of the image border are excluded, since a vehicle
  straddling the frame edge has truncated pixel area.
\item \emph{Precision} --- for every annotated box, whether at least
  $\tau_{\mathrm{phantom}} = \SI{10}{\percent}$ of the pixels it encloses are
  vehicle-class.
\end{itemize}
The $A_{\min}$ cut is set where the audit's own sensitivity and the
annotator's $\geq\!5$-point criterion begin to overlap, so that missed
detections are not conflated with two different filter regimes. The
\SI{10}{\percent} precision cut is deliberately permissive because an
axis-aligned box drawn around a rotated 3D box necessarily encloses
background: the worst case for a $45^{\circ}$-rotated rectangle is twice the
silhouette area, and audited boxes have a median vehicle-pixel coverage of
$0.58$ (p25 $= 0.47$). A box below \SI{10}{\percent} coverage cannot be
explained by that geometric excess and is essentially in the wrong place.

\paragraph{Sensitivity.}
Headline numbers are reported at the pre-registered
$(A_{\min}, \tau_{\mathrm{match}}, \tau_{\mathrm{phantom}})
= (100\,\mathrm{px}^{2}, 0.50, 0.10)$. Across the full grid
$A_{\min} \in \{50, 100, 150, 250, 400\}$\,px$^{2}$,
$\tau_{\mathrm{match}} \in \{0.35, 0.50, 0.65\}$ and
$\tau_{\mathrm{phantom}} \in \{0.05, 0.10, 0.15\}$, recall ranges from
\SI{79.2}{\percent} ($A_{\min} = 50$) to \SI{98.0}{\percent}
($A_{\min} = 400$) and varies by under two percentage points across
$\tau_{\mathrm{match}}$ at any fixed $A_{\min}$; precision ranges from
\SI{92.5}{\percent} ($\tau_{\mathrm{phantom}} = 0.15$) to \SI{96.7}{\percent}
($\tau_{\mathrm{phantom}} = 0.05$). The pre-registered values sit mid-grid.

\paragraph{Recall by distance.}
Recall is dominated by vehicle distance: a distant vehicle subtends few LiDAR
rays, making the $\geq\!5$-point criterion hard to satisfy. Binned by blob
area (an area-calibrated distance proxy), recall is \SI{99.9}{\percent} below
${\approx}\SI{20}{m}$, \SI{99.3}{\percent} at ${\approx}$\SIrange{20}{40}{m},
\SI{91.4}{\percent} at ${\approx}$\SIrange{40}{70}{m} and \SI{72.0}{\percent}
at ${\approx}$\SIrange{70}{100}{m}, for a weighted average of
\SI{87.8}{\percent}. The drop past ${\approx}\SI{40}{m}$ therefore reflects
the LiDAR's geometric point-density limit rather than an annotation defect;
Figure~\ref{fig:supp_failures}(a) shows a representative case.

\paragraph{False-positive taxonomy.}
Of the 13\,319 audited boxes, 765 fall below \SI{10}{\percent} vehicle-pixel
coverage. Of those, 112 ($\approx\SI{0.8}{\percent}$ of all annotated boxes)
are near-plane projection artefacts: the open-source projection routine does
not clip the 3D box against the camera near plane, so a vehicle passing very
close to the ego produces a 2D box spanning most of the image
(Figure~\ref{fig:supp_failures}(b)). They are reliably identified by the
heuristic \emph{box area $\geq\SI{50}{\percent}$ of image area \textbf{or}
box touches $\geq 3$ image edges} (see also \S\ref{sec:supp_projection}).
The remaining 653 ($\SI{85}{\percent}$ of the 765) are true LiDAR-clipping
phantoms: the LiDAR cone, originating from a different point than the camera,
clips through occluders (overpasses, fences, low walls) the camera cannot see
through (Figure~\ref{fig:supp_failures}(c)). The per-stratum pattern confirms
multi-level geometry as the dominant cause: Town04 (multi-level) has the
lowest precision (\SIrange{87}{89}{\percent}) and Town10 (flat dense urban)
the highest (\SI{99}{\percent}).

Because the near-plane artefacts originate in the projection routine rather
than in the visibility filter, we report the filter's precision with them
excluded (\SI{95.1}{\percent}) alongside the strict lower bound that counts
them as filter failures (\SI{94.3}{\percent}). Downstream users can remove
them at read time with the heuristic above.

\paragraph{Per-camera breakdown.}
The multi-camera extension shares one semantic LiDAR across four cameras, so
visibility can be misjudged for cameras far from the LiDAR.
\texttt{CAM\_FRONT} and \texttt{CAM\_BACK}, close to the LiDAR scan plane,
agree with the LiDAR-derived annotations within a fraction of a percentage
point (recall \SIrange{88.1}{90.3}{\percent}, precision excl.\ near-plane
\SIrange{95.7}{96.3}{\percent}). \texttt{CAM\_LEFT} and \texttt{CAM\_RIGHT}
drop $\sim$5 points in recall (84.8\,\%/85.0\,\%), and \texttt{CAM\_RIGHT}
a further 6 points in precision (89.3\,\%) -- the expected signature of the
multi-camera extension's geometric limitation.

\section{Experiment 2: full detection metrics}
\label{sec:supp_exp2}

The main paper (\S 4.2) reports mAP@0.5, precision, and recall per pipeline
and fog level. Table~\ref{tab:supp_exp2_full} gives the complete metrics,
adding mAP@.5:.95 and per-class AP (car, truck, large vehicle).

\paragraph{Training and validation settings.}
The YOLOv8-m detector is fine-tuned end-to-end from COCO-pretrained weights
for up to 75 epochs (early-stopping patience 19, batch size 48, image size
864\,px, SGD momentum 0.937, initial LR 0.01, weight decay $5\times10^{-4}$,
default Ultralytics augmentations). Detection is reported at the standard
Ultralytics validation settings (confidence threshold 0.001, NMS IoU 0.6).

\end{document}